\documentclass[sigconf,preprint]{acmart}

\usepackage{latexsym}
\usepackage{graphicx}
\usepackage{xcolor}
\usepackage{soul} 
\usepackage{pifont}
\usepackage[shortlabels]{enumitem}
\usepackage{booktabs}
\usepackage{multirow}
\usepackage{colortbl}
\usepackage{array}
\usepackage{tabularx}
\usepackage{pifont}
\usepackage{bm}
\usepackage{tabularray}
\usepackage{makecell}
\usepackage{stackengine}
\usepackage{tcolorbox}
\usepackage{threeparttable}
\usepackage{arydshln}
\tcbuselibrary{breakable}
\newsavebox{\measurebox}

\usepackage[dvipsnames]{xcolor}

\usepackage{hyperref}
\hypersetup{
    linkcolor=Red,       
    filecolor=Mulberry,   
    urlcolor=NavyBlue       
}

\usepackage[T1]{fontenc}
\usepackage[utf8]{inputenc}
\usepackage{microtype}
\usepackage{subfig}

\definecolor{myred}{RGB}{255,200,200}
\definecolor{myblue}{RGB}{200,220,255}
\definecolor{mygray}{RGB}{190,190,190}
\definecolor{method}{RGB}{217,217,252}

\newcommand{\cmark}{\ding{51}}

\usepackage{adjustbox}
\definecolor{oursrow}{rgb}{0.93,0.95,0.97}
\definecolor{oursrowdark}{rgb}{0.70,0.77,0.87}

\usepackage{calc}
\newcolumntype{C}[1]{>{\centering\arraybackslash}p{#1}}%
\newlength{\mycolwidth}
\newcolumntype{L}{>{\raggedright\arraybackslash}X}

\newcolumntype{T}{>{\fontsize{8.2}{9}\selectfont}c}
\newcolumntype{H}{>{\setbox0=\hbox\bgroup}c<{\egroup}@{}}

\AtBeginDocument{%
  }

\setcopyright{acmlicensed}
\copyrightyear{2027}
\acmYear{2027}
\acmDOI{XXXXXXX.XXXXXXX}

\begin{document}

%%
%% The "title" command has an optional parameter,
%% allowing the author to define a "short title" to be used in page headers.

\title{Self-Supervised Latent Action-Conditioned World Modeling for Time Series}

\author{Chia-Han Kuo$^*$}
\affiliation{
  \institution{National Yang Ming Chiao Tung University}
  \city{}
  \country{}}
\email{ruckuo.cs15@nycu.edu.tw}

\author{Chengshuai Zhao$^*$}
\affiliation{
  \institution{Arizona State University}
  \city{}
  \country{}}
\email{czhao93@asu.edu}

\author{Huan Liu}
\affiliation{
  \institution{Arizona State University}
  \city{}
  \country{}}
\email{huanliu@asu.edu}

\author{Lo Pang-Yun Ting}
\affiliation{
  \institution{National Yang Ming Chiao Tung University}
  \city{}
  \country{}}
\email{lpyting@nycu.edu.tw}




\begin{abstract}

Time-series data play an important role in various tasks, such as weather forecasting, web traffic analysis, and speech recognition. Recently, pioneering studies have increasingly adopted Joint-Embedding Predictive Architecture (JEPA) based encoders to learn time-series representations through latent-space prediction. However, they often overlook the underlying factors governing transitions between time-series observations. In this work, we hypothesize that \emph{explicitly modeling the information behind changes in time series can facilitate time-series modeling}. Building on this hypothesis, we propose \textbf{ACTS-WM}, a self-supervised latent action-conditioned world model framework for time-series analysis. First, we introduce a latent action modeling approach that infers the underlying information behind time-series transitions as \emph{latent actions}, without requiring explicit action annotations. We then develop a JEPA-based encoder and an action-conditioned world model to predict target time-series representations conditioned on the learned latent actions and task information. We evaluate ACTS-WM across 4 representative tasks on 22 datasets from diverse domains, covering univariate and multivariate forecasting and classification. Comprehensive experiments demonstrate that ACTS-WM can improve downstream performance over pretrained JEPA-based encoders by learning complementary representations.

\end{abstract}

\maketitle
\def\thefootnote{$*$}\footnotetext{Equal contribution.}

\section{Introduction}

\begin{figure}[t]
\graphicspath{{figs/}}
\begin{center}
\includegraphics[width=0.47\textwidth]{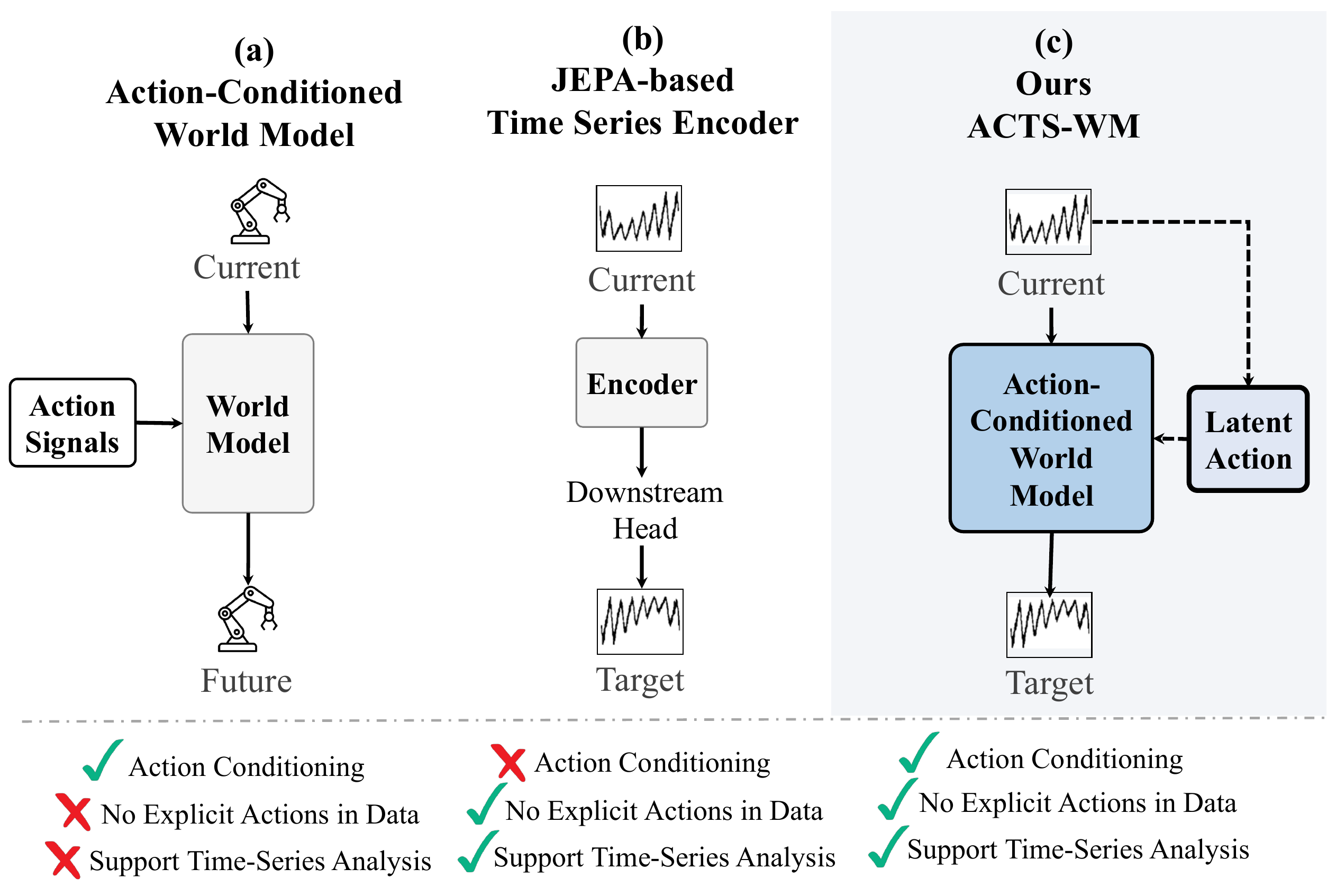}
\end{center}
\caption{Model comparison. (a) Conventional action-conditioned world models in robotics rely on explicit action signals. (b) JEPA-based time-series encoders learn representations without explicit action conditioning. (c) ACTS-WM learns latent actions from observational time series, enabling action-conditioned world modeling for time-series analysis without requiring explicit action signals.}
\label{fig:scenario}
\end{figure}

Time-series data play a pivotal role across modern data-driven systems, ranging from weather forecasting and energy monitoring to real-time speech recognition~\cite{fu2025on, zhou2021informer, 9414560}. In Web ecosystems, modern online services continuously generate massive volumes of time-series data, including Web traffic, user activity logs, and network monitoring signals~\cite{10.1145/3774904.3792374, 10.1145/3292500.3330680}, providing abundant resources for web knowledge discovery. Recently, time-series foundation models~\cite{goswami2024moment, cohen2024toto, liu2025moirai} have emerged as a state-of-the-art technique to leverage substantial time-series data. By embedding temporal data into a condensed latent space, they learn general-purpose representations supporting various tasks, including forecasting and classification.

To improve generalization across tasks and domains, recent studies have increasingly explored self-supervised representation learning for time series~\cite{fraikin2024t, nie2023a}. Among these approaches, Joint-Embedding Predictive Architectures (JEPA)~\cite{lecun2022path}, which learn representations by predicting target embeddings from contextual observations in latent space, demonstrate effectiveness across downstream time-series tasks~\cite{ennadir2024joint, he2026scjepa, dutta2026giving}.

Despite this progress, existing approaches primarily learn representations from observed temporal patterns without explicitly modeling the information underlying how time series change over time. Such information can characterize temporal transitions that are predictive of future dynamics while revealing invariant semantics shared across observations, thereby providing informative signals for both forecasting and classification. For example, pioneering studies in visual control and robotics have explored learning latent actions~\cite{schmidt2024learning, ye2025latent} from observed transitions. Inspired by this intuition, we hypothesize that \emph{explicitly modeling the information underlying temporal changes can provide richer supervisory signals and facilitate more effective time-series representation learning.}

However, modeling the information underlying temporal changes in time series presents several challenges. \textbf{\textit{\underline{(i)}}} Unlike conventional settings such as robotics~\cite{assran2025v, khorrambakht2026worldplanner, wu2023daydreamer, guo2026ctrl}, where an agent interacts with its environment through explicit control actions and observes the resulting state transitions, time-series data are typically collected as \emph{passive observations} without corresponding action signals, as shown in Figure~\ref{fig:scenario}. As a result, the latent factors that drive transitions between observations are not directly available and must instead be inferred from the observed temporal evolution itself. \textbf{\textit{\underline{(ii)}}} Further, multivariate time series can contain varying numbers of channels across datasets, making it difficult to model temporal transitions within a unified framework. These challenges call for a general mechanism that can infer latent transition factors from observed changes while remaining robust to heterogeneous and dataset-specific channel configurations.

To address these challenges, we propose \textbf{ACTS-WM} (\textbf{\ul{A}}c\-tion-\textbf{\ul{C}}on\-di\-tioned \textbf{\ul{T}}ime-\textbf{\ul{S}}e\-ries \textbf{\ul{W}}orld \textbf{\ul{M}}od\-el\-ing), a self-supervised latent action-conditioned world modeling framework for time-series analysis. ACTS-WM explicitly learns latent actions from observed time series and then leverages an action-conditioned world model for downstream forecasting and classification. Specifically, ACTS-WM consists of two stages. First, we introduce self-supervised latent action modeling to learn discrete latent actions from observed time-series transitions without requiring action annotations. We further learn to infer these actions from observed time series alone, enabling their use when target time series are unavailable. Second, we pretrain a JEPA-based time-series encoder to learn informative latent representations of time-series observations. We then develop an action-conditioned world model that incorporates the learned latent actions to model transitions between observations in the encoder's latent representation space. Through task-conditioned modeling, ACTS-WM is able to provide complementary latent representations to support downstream forecasting and classification within a unified framework.

\noindent\textbf{Key Takeaways.} Our contributions are summarized as follows:

\begin{itemize}[leftmargin=*]
    \item We propose a novel perspective for time-series representation learning. We argue that the latent factors underlying transitions between observations provide valuable supervisory signals for learning transferable representations.
    \item Building on the hypothesis, we develop a task-conditioned latent action modeling approach that captures temporal evolution and variations among time series of the same class.
    \item We propose ACTS-WM. To the best of our knowledge, it is the first self-supervised latent action-conditioned world-modeling framework tailored for time-series analysis.
    \item To validate the hypothesis, we conduct extensive experiments across univariate and multivariate forecasting and classification tasks on 22 time-series datasets from diverse domains. The results demonstrate the effectiveness of ACTS-WM in providing complementary representations, while further ablation studies and qualitative analyses reveal meaningful transition patterns captured by the learned latent actions.
\end{itemize}

\section{Related Work}

\begin{figure*}[t]
\centering
\includegraphics[width=0.97\textwidth]{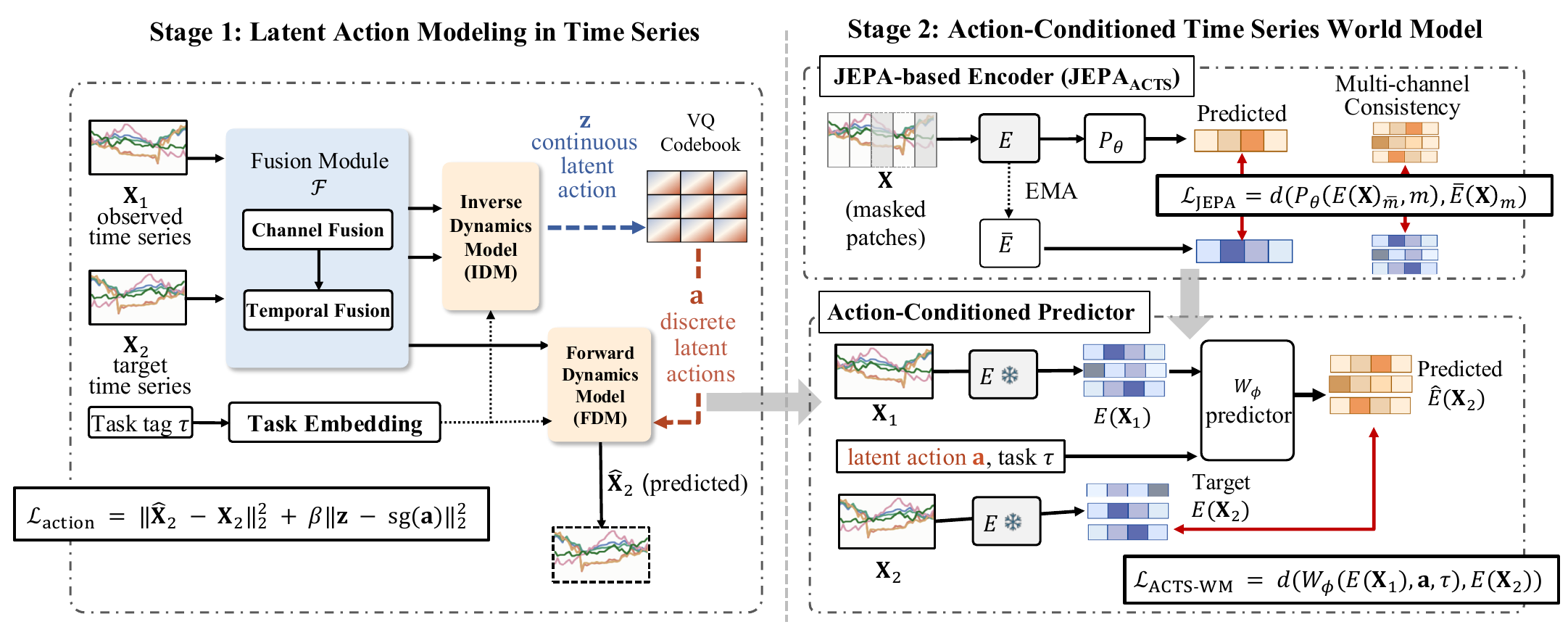}
\vspace{-1em}
\caption{Overview of ACTS-WM.}
\label{fig:framework}
\end{figure*}

\subsection{Time Series Analysis}

For time-series analysis models, we explore time series foundation models and JEPA-based time-series models. Recent studies have explored various approaches to time-series foundation models.
These studies introduce novel architectures that achieve remarkable performance on multi-domain benchmarks. MOMENT~\cite{goswami2024moment} and UniTS~\cite{NEURIPS2024_fe248e22} learn expressive representations that can be directly adapted to various downstream tasks. Toto~\cite{cohen2024toto}, Moirai~\cite{pmlr-v235-woo24a,liu2025moirai, liu2025moiraimoe}, and Time-MoE~\cite{shi2025timemoe}, also learn expressive representations but focus on forecasting ability.

In parallel, Joint-Embedding Predictive Architectures (JEPA)~\cite{lecun2022path} have gained increasing attention in time-series analysis. TS-JEPA~\cite{ennadir2024joint} and CHARM~\cite{dutta2026giving} adapt the JEPA architecture to various time-series analysis tasks. HEPA~\cite{petersen2026hepa} and SC-JEPA~\cite{he2026scjepa} learn to predict future latent representations and apply them to rare-event prediction and anomaly detection. LaT-PFN~\cite{verdenius2024latpfnjointembeddingpredictive} further combines JEPA with zero-shot forecasting capabilities. Beyond representation learning, JEPA-based time-series world models have explored incorporating additional action signals, such as mechanical states and control inputs~\cite{qi2026staticforecastingunleashingpower,shaj2023multi} or external information from other modalities~\cite{dai2026pythia}.

\subsection{World Models}

World models aim to capture the dynamics of the physical world and support a wide range of downstream tasks. Dreamer 4~\cite{hafner2025trainingagentsinsidescalable} learns a scalable latent dynamics model, while DIAMOND~\cite{NEURIPS2024_6bdde037} adopts a diffusion model as a pixel-space world model. Both then train policies via reinforcement learning in imagination.
Extending the JEPA framework, V-JEPA 2-AC~\cite{assran2025v} plans in latent space to control robot arms for picking and placing objects given a goal image. H-JEPA~\cite{zhang2026h} further introduces a hierarchy of latent world models that enables long-horizon planning at multiple timescales.
Video-generation-based world models such as COSMOS~\cite{agarwal2026cosmos} and Genie~\cite{bruce2024genie} generate physically plausible, action-controllable videos.
Some works aim to combine policy models and world models. PaW~\cite{lu2026policy} employs a single pretrained large language model (LLM) to jointly train a policy and a world model, while CoLA-World~\cite{wang2025co} jointly trains a latent action model (LAM) with a pretrained video world model, allowing the two to co-evolve.

\subsection{Latent Action Models}
\label{subsec:lam}
To learn dynamics from observational data without explicit action labels, recent studies~\cite{schmidt2024learning, ye2025latent, gao2025adaworld, wang2025co} have explored latent action learning to infer action representations directly from observed transitions. These studies employ inverse dynamics models (IDMs) or latent action encoders to infer latent actions from observed transitions, and forward dynamics models (FDMs) or latent action decoders to predict subsequent observations conditioned on the inferred actions. LAPO~\cite{schmidt2024learning}, LAPA~\cite{ye2025latent} and CoLA-World~\cite{wang2025co} further quantize the learned latent actions into a discrete action space.

Despite these advances, most time-series models do not explicitly model the underlying information governing transitions between observations. Existing world models primarily focus on visual and language domains, while time-series world models often rely on additional action signals, such as system states or control inputs. Although latent action models enable modeling transitions without explicit action labels, they have mainly been explored in video-based settings. 
In our work, ACTS-WM aims to learn latent actions directly from time-series observations and uses them to condition world modeling without requiring action annotations.

\section{Preliminaries and Problem Definition}

\subsection{Problem Formulation}
\label{sec:formulation}

Let $\mathbf{X} \in \mathbb{R}^{C \times T}$ denote a time series with $C$ channels and $T$ time steps, where $C > 1$ corresponds to the multivariate setting. Given time-series observations collected from multiple domains, we define the pretraining dataset as: $\mathcal{D}=\{(\mathbf{X}_1^{(i)},\mathbf{X}_2^{(i)},\tau^{(i)})\}_{i=1}^{N}$, where $\mathbf{X}_1 \in \mathbb{R}^{C \times T_1}$ denotes the observed time series, and $\mathbf{X}_2 \in \mathbb{R}^{C \times T_2}$ denotes the target time series. For instance, in forecasting tasks, $\mathbf{X}_2$ represents the subsequent temporal segment following $\mathbf{X}_1$. The task tag $\tau$ indicates whether the observation pair ($\mathbf{X}_1$, $\mathbf{X}_2$) corresponds to a forecasting task with a specified prediction horizon or a classification task.

Given the pretraining dataset $\mathcal{D}$, ACTS-WM aims to learn latent actions that capture the underlying transition information between time-series observations without requiring explicit action annotations. At the evaluation phase, given an unseen time series $\mathbf{X}'_1$ from either a pretraining domain or a previously unseen domain, along with a task tag $\tau$, ACTS-WM infers a latent action and uses it to predict the latent representations of the target time series. These predicted representations are subsequently used for downstream forecasting and classification.

\subsection{Inverse and Forward Dynamics Models}
\label{sec:prelim_idm}

Inverse and Forward dynamics models are two main components in recent latent action models to learn latent actions from video data, as discussed in Sec.~\ref{subsec:lam}. The inverse dynamics model (IDM) $p_{IDM}$ infers latent actions $z_t$ from past observations $\{o_{t-k}, ..., o_{t}\}$ and future observation $o_{t+1}$, while the forward dynamics model (FDM) $p_{FDM}$ reconstructs the future observation $o_{t+1}$ given the past observations and a latent action obtained from IDM. These processes are formulated as:

\[
z_t \sim p_{\mathrm{IDM}}(\cdot \mid o_{t-k},\ldots,o_t,o_{t+1}),
\]
\[
\hat{o}_{t+1} \sim p_{\mathrm{FDM}}(\cdot \mid o_{t-k},\ldots,o_t,z_t).
\]

In studies~\cite{schmidt2024learning, ye2025latent, wang2025co}, vector quantization~\cite{gray1984vector} is further applied to latent actions before passing them to the FDM, serving as an information bottleneck that prevents the IDM from simply copying future observations into the action representation.

In our work, we build upon these approaches and extend latent action modeling to time-series data.

\section{Methodology}
\label{sec:method}

Our \textbf{ACTS-WM} framework consist of two stages, as shown in Figure~\ref{fig:framework}: (1) \emph{Latent Action Modeling} learns discrete latent actions from observed time series
(Sec.~\ref{sec:latent_action}), and (2) \emph{Action-Conditioned Time Series World Model} uses the learned actions and a JEPA-based encoder to predict representations of target time series (Sec.~\ref{sec:wm}).

\subsection{Latent Action Modeling in Time Series}
\label{sec:latent_action}

In our work, the underlying information that explains the transition between observed and target time-series observations is regarded as a \emph{latent action}.
Building upon the latent action learning framework described in Sec.~\ref{sec:prelim_idm}, we introduce several designs to learn latent actions from time-series data.

Given an observation pair $(\mathbf X_1,\mathbf X_2,\tau)\in\mathcal D$, we first encode the time-series observations $(\mathbf X_1,\mathbf X_2)$ and task information $\tau$ separately. These representations are then incorporated into the inverse and forward dynamics models to learn latent actions.

\subsubsection{Channel and Temporal Fusions}
\label{subsubsec:fusion}

To accommodate varying numbers of channels and sequence lengths in time-series data, we develop a fusion module $\mathcal{F}$ consisting of two components: Channel Fusion and Temporal Fusion. For \textbf{Channel Fusion}, we first fuse features from each channel independently using shared convolutional layers, followed by multi-head self-attention and learnable-query attention pooling to capture and aggregate cross-channel dependencies. \textbf{Temporal Fusion} then employs dilated temporal convolutional networks (TCNs)~\cite{bai2018empirical, lin2021temporal} to capture multi-scale temporal patterns. Given a time series $\mathbf{X} \in \mathbb{R}^{C \times T}$, its encoding process is formulated as:

\begin{equation}
\mathcal{F}(\mathbf X)
= \operatorname{Fusion}_{\text{temporal}}
\left(\operatorname{Fusion}_{\text{channel}}(\mathbf X)\right)
\in\mathbb R^{d_{\text{fuse}}},
\end{equation}
where $d_{\text{fuse}}$ is the dimension of the resulting representation. The extracted features are subsequently used for latent action modeling.

\subsubsection{Task Embedding}
\label{subsubsec:task_embed}

To distinguish different types of transitions, including forecasting with different prediction horizons and classification, we introduce learnable \textbf{Task Embeddings}. Given a task tag $\tau$, its embedding is obtained as:

\begin{equation}
    \mathbf e_\tau=\mathbf E\cdot\mathbf I_\tau,
\end{equation}
where $\mathbf I_\tau$ denotes the one-hot vector corresponding to the task tag $\tau$, and $\mathbf E$ is a learnable embedding table with embedding dimension $d_\tau$. Each task tag is assigned a unique index to construct its one-hot vector. The embedding table is optimized jointly with the model parameters through the training objective, allowing the model to learn task-specific representations.

\subsubsection{Inverse and Forward Dynamics Models}
\label{sec:idm}
Building upon the preceding designs, we redesign the inverse dynamics model (IDM) $f_{\mathrm{IDM}}$ to infer a continuous latent action from an observation pair $(\mathbf X_1,\mathbf X_2,\tau)\in\mathcal D$. Since $\mathbf X_1$ and $\mathbf X_2$ may have different sequence lengths, we employ the fusion module $\mathcal F$ (Sec.~\ref{subsubsec:fusion}) to map them into fixed-dimensional representations of the same dimension. These representations are then combined with the task embedding $\mathbf e_\tau$ (Sec.~\ref{subsubsec:task_embed}) to infer the latent action $\mathbf z$:

\begin{equation}
\label{eq:idm}
\begin{aligned}
&\hspace*{0cm}
\mathbf z=f_{IDM}(\mathcal F(\mathbf X_1),\mathcal F(\mathbf X_2),e_\tau)
\\
&\hspace*{0.2cm}
=\mathbf W_2 \left[\operatorname{ReLU}\left(\mathbf W_1[\mathcal F(\mathbf X_1);\mathcal F(\mathbf X_2)]\right);\mathbf e_\tau\right], 
\end{aligned}
\end{equation}
where $\mathbf W_1$ and $\mathbf W_2$ are learnable projection matrices, and $[\cdot;\cdot]$ denotes concatenation. The inferred action $\mathbf z$ represents the transition of the observation pair and is shared across all channels. 

Following ~\cite{schmidt2024learning, ye2025latent, wang2025co}, the continuous action $\mathbf z$ is then quantized into a discrete latent action $\mathbf a=VQ(\mathbf z)$ using vector quantization (details in Sec.~\ref{sec:vq}). Correspondingly, we redesign the forward dynamics model (FDM) $f_{\mathrm{FDM}}$ to reconstruct the target observation $\mathbf X_2$ from $\mathbf X_1$ and the quantized latent action $\mathbf a$, formulated as:

\vspace{-1em}
\begin{equation}
    \hat{\mathbf X}_2=f_{\mathrm{FDM}}(\mathcal F(\mathbf X_1), \mathbf a, \mathbf e_\tau),
\end{equation}
where $\hat{\mathbf X}_2$ denotes the reconstructed target observation. Specifically, we employ a one-dimensional U-Net~\cite{ronneberger2015u} with shared weights across channels to accommodate varying numbers of channels. The FDM incorporates the quantized latent action $\mathbf a$, the fused representation $\mathcal F(\mathbf X_1)$, and the task embedding $\mathbf e_\tau$ as conditioning information.

Finally, the IDM $f_{\mathrm{IDM}}$ and FDM $f_{\mathrm{FDM}}$ are jointly trained to minimize the reconstruction loss for time series $\mathbf X_2$ and a quantization loss that encourages the continuous latent action to remain close to its quantized representation using the following objective:

\begin{equation}
\label{eq:loss_action}
\begin{aligned}
&\hspace*{0cm}
\mathcal L_{\mathrm{action}}
= \left\|f_{FDM}(F(\mathbf X_1),F(\mathbf X_2),e_\tau)-\mathbf X_2\right\|_2^2
\\
&\hspace*{1.2cm}
+\beta\cdot\left\|f_{IDM}(\mathcal F(\mathbf X_1),\mathcal F(\mathbf X_2),e_\tau)-\operatorname{sg}(\mathbf a)\right\|_2^2,
\end{aligned}
\end{equation}
where the first term measures the reconstruction error, and the second term denotes the quantization loss weighted by $\beta$. Here, $\operatorname{sg}(\cdot)$ denotes the stop-gradient operator. This objective encourages the IDM to encode information relevant to the transition toward $\mathbf X_2$ in the latent action.

\subsubsection{Latent Action Selection}
\label{sec:vq}
We describe the latent action quantization process and how latent actions are obtained at the evaluation phase by using the latent policy.

\noindent\textbf{Latent Action Quantization.} Following previous studies~\cite{schmidt2024learning, ye2025latent, wang2025co}, vector quantization is applied to the continuous latent action $\mathbf z$ to introduce an information bottleneck. Specifically, a codebook is constructed consisting of $K$ code vectors, each with dimension $d_{\text{code}}$, and the continuous latent action $\mathbf z$ is divided into $M$ slots. Each slot is quantized by selecting its nearest code vector from the codebook, and the selected vectors are concatenated to form the quantized latent action $\mathbf a=VQ(\mathbf z)$. The codebook vectors are updated using an exponential moving average, while $K$ determines the number of available codes and $M$ determines the number of code selections per action.

\noindent\textbf{Latent Policy.} Following LAPO~\cite{schmidt2024learning}, since $\mathbf X_2$ is unavailable at the evaluation phase, we train a latent policy $\pi(\mathbf X_1, \tau)$ to infer latent actions from observed time series $\mathbf X_1$ and the task tag $\tau$ alone. With the pretrained IDM frozen, the latent policy is trained to regress the continuous latent actions produced by the IDM from observation pairs. At the evaluation phase, the policy's output is quantized using the learned codebook to obtain the latent action.

\subsection{Action-Conditioned Time-Series World Model}
\label{sec:wm}

After learning latent actions from time-series observations, we develop an action-conditioned world model by adapting a JEPA-based encoder for time-series representation learning and training a predictor to infer representations of target observations.

\subsubsection{JEPA-based Encoder}
\label{sec:vjepa}

We build upon V-JEPA 2~\cite{assran2025v}, which learns predictive spatiotemporal representations from video data, making it suitable for time-series modeling. To adapt it to time-series data, we introduce one-dimensional temporal patches and channel-independent encoding with shared weights, allowing the model to handle varying numbers of channels. The adapted encoder $E$ maps a time series $\mathbf X\in\mathbb R^{C\times T}$ to its latent representation: $E:\mathbb R^{C\times T}\rightarrow\mathbb R^{C\times N\times d_{\text{jepa}}},
$ where $N$ is the number of temporal patches and $d_{\text{jepa}}$ is the representation dimension. During pretraining, we employ interior-block and suffix masking and retain the exponential moving average (EMA) target encoder of V-JEPA 2, which provides stable representations for masked prediction.

To further account for dependencies across channels, we introduce a multi-channel consistency term. Given the predicted and target representations $\hat{\mathbf E}$ and $\mathbf E$, respectively, we define the multi-channel distance function as:

\vspace{-1em}
\begin{equation}
\label{eq:channel_consist}
    d(\hat{\mathbf E},\mathbf E) = \frac{1}{C}
    \left[\sum_{c=1}^{C}
    \left\|\hat{\mathbf E}_c-\mathbf E_c\right\|_1
    +
    \lambda
    \left\|
    \tfrac{1}{C}\textstyle\sum_{c=1}^{C}\hat{\mathbf E}_c
    -
    \tfrac{1}{C}\textstyle\sum_{c=1}^{C}\mathbf E_c
    \right\|_1
    \right],
\end{equation}
where $\hat{\mathbf E}_c$ and $\mathbf E_c$ denote the predicted and target representations of channel $c$ at the masked positions, respectively, and $\lambda$ weights the multi-channel consistency term.

Given a binary temporal mask $\mathbf m$ indicating the masked patch positions and its complement $\bar{\mathbf m}$ (unmasked positions), our JEPA-based encoder $E$ and predictor $P_\theta$ are jointly trained to minimize:

\begin{equation}
\label{eq:loss_jepa}
    \mathcal L_{\mathrm{JEPA}}
    =
    d\left(
    P_\theta(E(\mathbf X)_{\bar{\mathbf m}},\mathbf m),
    \bar E(\mathbf X)_{\mathbf m}
    \right),
\end{equation}
where $E(\mathbf X)_{\bar{\mathbf m}}$ denotes the representations encoded from the unmasked patches, $P_\theta$ predicts the representations at the masked positions, and $\bar E(\mathbf X)_{\mathbf m}$ denotes the corresponding target representations obtained from the EMA target encoder. 

By optimizing Eq.~\ref{eq:loss_jepa}, the JEPA-based encoder $E$ learns latent representations of time-series observations for subsequent world modeling.

\subsubsection{Action-Conditioned Predictor}
\label{sec:ac_predictor}

V-JEPA 2-AC~\cite{assran2025v} is a latent action-conditioned world model post-trained from V-JEPA 2. Building upon this approach, we adapt its action-conditioned predictor $W_\phi$ (with parameters $\phi$) to model transitions between time-series representations. Given the JEPA-based encoder $E$, the quantized latent action $\mathbf a$, and the task tag $\tau$, the predictor estimates the representation of target observation $\mathbf X_2$ as: $W_\phi(E(\mathbf X_1),\mathbf a,\tau)$. We then optimize the action-conditioned predictor $W_\phi$ using the multi-channel distance function $d$ introduced in Eq.~\ref{eq:channel_consist}:

\begin{equation}
\mathcal L_{\mathrm{ACTS-WM}}=d\left(
W_\phi(E(\mathbf X_1),\mathbf a,\tau),
(E(\mathbf X_2))
\right).
\end{equation}

Overall, the training objectives of ACTS-WM are as follows:

\begin{equation}
\label{eq:loss_wm_total}
\begin{aligned}
\min_{\phi}\quad & \mathcal L_{\mathrm{ACTS-WM}}\\
\mathrm{s.t.}\quad
& \theta_{\mathrm{action}}^{*}
=\arg\min_{\theta_{\mathrm{action}}}
\mathcal L_{\mathrm{action}},\\
& \theta_{\mathrm{JEPA}}^{*}
=\arg\min_{\theta_{\mathrm{JEPA}}}
\mathcal L_{\mathrm{JEPA}},
\end{aligned}
\end{equation}
where $\theta_{\mathrm{action}}$ and $\theta_{\mathrm{JEPA}}$ are parameters of latent action modeling and JEPA pretraining, respectively.

Finally, by optimizing Eq.~\ref{eq:loss_wm_total}, ACTS-WM learns to predict target representations for time-series observations across different domains, sequence lengths, and numbers of channels, supporting different downstream tasks. While ACTS-WM is developed based on V-JEPA 2, the framework can also be integrated with other JEPA-based encoders and general time-series representation models.
\section{Experiments}

\subsection{Experimental Setup}

\subsubsection{Dataset.}

We collect univariate and multivariate time-series datasets from various sources, including the UEA~\cite{bagnall2018uea} and UCR~\cite{8894743} archives, forecasting benchmarks from Autoformer~\cite{NEURIPS2021_bcc0d400}, WebTraffic~\cite{ICLR2024_fb9d01fb}, and WikiTraffic~\cite{godahewa2021monash, godahewa_2022_7371038}. We use 21 datasets for ACTS-WM pretraining and 22 for downstream evaluation. Dataset details and usage are summarized in Table~\ref{tb:datasets}, while preprocessing and data splits are described in Appendix~\ref{appendix:data_settings}.

\subsubsection{Baselines.}
We compare ACTS-WM with publicly available JEPA-based time-series encoders: \textbf{TS-JEPA}~\cite{ennadir2024joint} and \textbf{SC-JEPA}~\cite{he2026scjepa}. We also evaluate our JEPA-based encoder (Sec.~\ref{sec:vjepa}), adapted from V-JEPA 2 to time series, denoted as ``\textbf{JEPA$_{\text{ACTS}}$}''.

For downstream evaluation, we employ task-specific heads for all methods: \emph{Linear}, \emph{Ridge}, and \emph{NLH} (non-linear head) for forecasting, and \emph{Linear}, \emph{MLP}, and \emph{SVM} for classification. Using these heads, we evaluate three representation settings for each JEPA-based encoder:
\begin{itemize}[leftmargin=*]
    \item ``\textbf{Base}'' uses only the pretrained encoder's representations.
    \item ``\textbf{Pred. Only}'' uses only the representations of the target time series predicted by ACTS-WM.
    \item ``\textbf{Obs.}+\textbf{Pred.}'' concatenates the representation of the observed time series from the JEPA-based encoder and the representation of the target time series by ACTS-WM.
\end{itemize}
Since TS-JEPA supports only univariate inputs, we evaluate it only on univariate tasks. Detailed settings for baselines and downstream heads are provided in Appendix~\ref{appendix:model_settings}.

\begin{table*}[t]
\small
\centering
\caption{Time-series forecasting results averaged across three runs. Datasets marked with $\ddagger$ are not included in pre-training ACTS-WM. The best result is in \textbf{bold}, and $^{\star}$ marks results with ACTS-WM that outperform the original base models. The win rate is the percentage of datasets where ACTS-WM outperforms the base model. Win rates above 50\% are shaded.}
\label{tb:jepa_forecast}
\setlength{\tabcolsep}{5.6pt} 
\renewcommand{\arraystretch}{0.7}
\begin{adjustbox}{max width=\textwidth}
\begin{tabular}{lr*{9}{c}}
\toprule
\multirow{2}{*}{\textbf{Dataset}} & \multirow{2}{*}{\textbf{Metric}}
& \multicolumn{3}{c}{Base}
& \multicolumn{3}{c}{+ \textbf{ACTS-WM} (Pred. Only)}
& \multicolumn{3}{c}{+ \textbf{ACTS-WM} (Obs.+Pred.)} \\
\cmidrule(lr){3-5} \cmidrule(lr){6-8} \cmidrule(lr){9-11}
& & JEPA$_{\text{ACTS}}$ & TS-JEPA & SC-JEPA & JEPA$_{\text{ACTS}}$ & TS-JEPA & SC-JEPA & JEPA$_{\text{ACTS}}$ & TS-JEPA & SC-JEPA \\
\midrule[0.6pt]
\multicolumn{11}{c}{\textit{\textbf{Univariate Time Series Forecasting}}} \\
\midrule[0.6pt]
 & MSE & 1.167 & 1.047 & 1.413 & 1.063$^{\star}$ & \textbf{0.972$^{\star}$} & 1.363$^{\star}$ & 1.164$^{\star}$ & 1.040$^{\star}$ & 1.906  \\
\multirow{-2}{*}{WikiTraffic$\ddagger$} & MAE & 0.705 & 0.664 & 0.807 & 0.652$^{\star}$ & \textbf{0.610$^{\star}$} & 0.773$^{\star}$ & 0.704$^{\star}$ & 0.677 & 0.981  \\
\rowcolor{gray!10}  & MSE & 0.007 & 0.002 & 0.003 & 0.003$^{\star}$ & \textbf{0.001$^{\star}$} & 0.003 & 0.007 & 0.002 & 0.003  \\
\rowcolor{gray!10} \multirow{-2}{*}{Weather} & MAE & 0.045 & 0.031 & 0.038 & 0.042$^{\star}$ & \textbf{0.025$^{\star}$} & 0.038 & 0.046 & 0.028$^{\star}$ & 0.044  \\
 & MSE & 0.100 & 0.097 & 1.631 & \textbf{0.084$^{\star}$} & 0.157 & 0.728$^{\star}$ & 0.086$^{\star}$ & 0.091$^{\star}$ & 1.739  \\
\multirow{-2}{*}{ETTh1} & MAE & 0.248 & 0.240 & 1.142 & \textbf{0.220$^{\star}$} & 0.310 & 0.710$^{\star}$ & 0.227$^{\star}$ & 0.234$^{\star}$ & 1.093$^{\star}$  \\
\rowcolor{gray!10}  & MSE & 0.478 & 0.402 & 1.478 & \textbf{0.351$^{\star}$} & \textbf{0.351$^{\star}$} & 1.358$^{\star}$ & 0.444$^{\star}$ & 0.396$^{\star}$ & 2.140  \\
\rowcolor{gray!10} \multirow{-2}{*}{ETTh2} & MAE & 0.516 & 0.508 & 1.018 & 0.467$^{\star}$ & \textbf{0.456$^{\star}$} & 0.985$^{\star}$ & 0.497$^{\star}$ & 0.500$^{\star}$ & 1.197  \\
 & MSE & 0.044 & 0.045 & 1.312 & \textbf{0.035$^{\star}$} & 0.045 & 0.793$^{\star}$ & 0.043$^{\star}$ & 0.054 & 0.775$^{\star}$  \\
\multirow{-2}{*}{ETTm1} & MAE & 0.158 & 0.160 & 0.990 & \textbf{0.135$^{\star}$} & 0.158$^{\star}$ & 0.753$^{\star}$ & 0.155$^{\star}$ & 0.177 & 0.736$^{\star}$  \\
\rowcolor{gray!10}  & MSE & 0.184 & 0.197 & 1.602 & 0.138$^{\star}$ & 0.184$^{\star}$ & 1.673 & 0.175$^{\star}$ & \textbf{0.133$^{\star}$} & 1.073$^{\star}$  \\
\rowcolor{gray!10} \multirow{-2}{*}{ETTm2} & MAE & 0.332 & 0.352 & 1.063 & 0.286$^{\star}$ & 0.332$^{\star}$ & 1.043$^{\star}$ & 0.324$^{\star}$ & \textbf{0.270$^{\star}$} & 0.865$^{\star}$  \\
\midrule[0.3pt]
\midrule[0.3pt]
\textit{Win rate vs.\ Base} & MSE & -- & -- & -- & \cellcolor{oursrowdark}100\% & \cellcolor{oursrowdark}67\% & \cellcolor{oursrowdark}67\% & \cellcolor{oursrowdark}83\% & \cellcolor{oursrowdark}67\% & 33\%    \\
 & MAE & -- & -- & -- & \cellcolor{oursrowdark}100\% & \cellcolor{oursrowdark}83\% & \cellcolor{oursrowdark}83\% & \cellcolor{oursrowdark}83\% & \cellcolor{oursrowdark}67\% & 50\%    \\
\midrule[0.6pt]
 \multicolumn{11}{c}{\textit{\textbf{Multivariate Time Series Forecasting}}} \\
\midrule[0.6pt]
 & MSE & 0.257 & -- & 0.590 & \textbf{0.238$^{\star}$} & -- & 0.471$^{\star}$ & 0.259 & -- & 0.514$^{\star}$  \\
\multirow{-2}{*}{Weather} & MAE & \textbf{0.225} & -- & 0.500 & 0.238 & -- & 0.418$^{\star}$ & \textbf{0.225} & -- & 0.456$^{\star}$  \\
\rowcolor{gray!10}  & MSE & 1.218 & -- & 1.321 & \textbf{0.678$^{\star}$} & -- & 1.129$^{\star}$ & 1.196$^{\star}$ & -- & 1.209$^{\star}$  \\
\rowcolor{gray!10} \multirow{-2}{*}{ETTh1} & MAE & 0.742 & -- & 0.888 & \textbf{0.557$^{\star}$} & -- & 0.825$^{\star}$ & 0.737$^{\star}$ & -- & 0.855$^{\star}$  \\
 & MSE & 0.366 & -- & 1.890 & \textbf{0.276$^{\star}$} & -- & 1.395$^{\star}$ & 0.297$^{\star}$ & -- & 2.084  \\
\multirow{-2}{*}{ETTh2} & MAE & 0.423 & -- & 1.080 & \textbf{0.376$^{\star}$} & -- & 0.922$^{\star}$ & 0.396$^{\star}$ & -- & 1.142  \\
\rowcolor{gray!10}  & MSE & 0.597 & -- & 1.434 & \textbf{0.489$^{\star}$} & -- & 1.105$^{\star}$ & 0.564$^{\star}$ & -- & 1.599  \\
\rowcolor{gray!10} \multirow{-2}{*}{ETTm1} & MAE & 0.513 & -- & 0.924 & \textbf{0.456$^{\star}$} & -- & 0.805$^{\star}$ & 0.503$^{\star}$ & -- & 0.987  \\
 & MSE & 0.161 & -- & 1.913 & \textbf{0.152$^{\star}$} & -- & 1.596$^{\star}$ & 0.156$^{\star}$ & -- & 1.740$^{\star}$  \\
\multirow{-2}{*}{ETTm2} & MAE & 0.281 & -- & 1.090 & \textbf{0.270$^{\star}$} & -- & 0.973$^{\star}$ & 0.276$^{\star}$ & -- & 1.031$^{\star}$  \\
\midrule[0.3pt]
\midrule[0.3pt]
\textit{Win rate vs.\ Base} & MSE & -- & -- & -- & \cellcolor{oursrowdark}100\% & -- & \cellcolor{oursrowdark}100\% & \cellcolor{oursrowdark}80\% & -- & \cellcolor{oursrowdark}60\%    \\
 & MAE & -- & -- & -- & \cellcolor{oursrowdark}80\% & -- & \cellcolor{oursrowdark}100\% & \cellcolor{oursrowdark}80\% & -- & \cellcolor{oursrowdark}60\%    \\
\bottomrule[1.3pt]
\end{tabular}
\end{adjustbox}
\end{table*}

\begin{table*}[t]
\centering
\small
\caption{Time-series classification results averaged over three runs. Datasets marked with $\ddagger$ are not included in pre-training ACTS-WM. The best result is in \textbf{bold}, and $^{\star}$ marks results with ACTS-WM that outperform the original base models. The win rate is the percentage of datasets where ACTS-WM outperforms the base model. Win rates above 50\% are shaded.}
\label{tb:jepa_classify}
\setlength{\tabcolsep}{4pt}
\renewcommand{\arraystretch}{0.7}
\begin{adjustbox}{max width=\textwidth}
\begin{tabular}{lr*{9}{c}}
\toprule
\multirow{2}{*}{\textbf{Dataset}} & \multirow{2}{*}{\textbf{Metric}}
& \multicolumn{3}{c}{Base}
& \multicolumn{3}{c}{+ \textbf{ACTS-WM} (Pred. Only)}
& \multicolumn{3}{c}{+ \textbf{ACTS-WM} (Obs.+Pred.)} \\
\cmidrule(lr){3-5} \cmidrule(lr){6-8} \cmidrule(lr){9-11}
& & JEPA$_{\text{ACTS}}$ & TS-JEPA & SC-JEPA & JEPA$_{\text{ACTS}}$ & TS-JEPA & SC-JEPA & JEPA$_{\text{ACTS}}$ & TS-JEPA & SC-JEPA \\
\midrule[0.6pt]
\multicolumn{11}{c}{\textit{\textbf{Univariate Time Series Classification}}} \\
\midrule[0.6pt]
WebTraffic$\ddagger$ & Acc. & 0.152 & 0.194 & \textbf{0.265} & 0.168$^{\star}$ & 0.158 & 0.215 & 0.158$^{\star}$ & 0.202$^{\star}$ & 0.213  \\
\rowcolor{gray!10} Earthquakes$\ddagger$ & Acc. & 0.715 & \textbf{0.748} & 0.667 & \textbf{0.748$^{\star}$} & \textbf{0.748} & 0.736$^{\star}$ & 0.739$^{\star}$ & \textbf{0.748} & 0.719$^{\star}$  \\
UWaveGestureLibraryX$\ddagger$ & Acc. & 0.633 & 0.635 & 0.588 & 0.663$^{\star}$ & \textbf{0.701$^{\star}$} & 0.563 & 0.653$^{\star}$ & 0.700$^{\star}$ & 0.577  \\
\rowcolor{gray!10} ShapesAll$\ddagger$ & Acc. & 0.474 & 0.652 & 0.622 & 0.521$^{\star}$ & 0.655$^{\star}$ & 0.637$^{\star}$ & 0.506$^{\star}$ & \textbf{0.661$^{\star}$} & 0.631$^{\star}$  \\
WormsTwoClass$\ddagger$ & Acc. & 0.541 & 0.545 & 0.532 & 0.554$^{\star}$ & 0.571$^{\star}$ & 0.623$^{\star}$ & 0.558$^{\star}$ & 0.545 & \textbf{0.636$^{\star}$}  \\
\rowcolor{gray!10} Computers$\ddagger$ & Acc. & 0.592 & 0.464 & 0.652 & 0.592 & 0.623$^{\star}$ & \textbf{0.692$^{\star}$} & 0.640$^{\star}$ & 0.500$^{\star}$ & \textbf{0.692$^{\star}$}  \\
InsectWingbeatSound$\ddagger$ & Acc. & 0.444 & 0.610 & \textbf{0.616} & 0.494$^{\star}$ & 0.493 & 0.522 & 0.478$^{\star}$ & 0.609 & 0.551  \\
\rowcolor{gray!10} EOGHorizontalSignal$\ddagger$ & Acc. & 0.355 & 0.382 & 0.180 & 0.328 & 0.388$^{\star}$ & 0.162 & 0.322 & \textbf{0.400$^{\star}$} & 0.181$^{\star}$  \\
EthanolLevel$\ddagger$ & Acc. & 0.360 & 0.298 & \textbf{0.457} & 0.246 & 0.270 & 0.379 & 0.396$^{\star}$ & 0.268 & 0.377  \\
\rowcolor{gray!10} SemgHandMovementCh2$\ddagger$ & Acc. & 0.347 & 0.496 & 0.331 & 0.484$^{\star}$ & \textbf{0.559$^{\star}$} & 0.220 & 0.520$^{\star}$ & 0.516$^{\star}$ & 0.239  \\
\midrule[0.3pt]
\midrule[0.3pt]
\textit{Win rate vs.\ Base} & Acc. & -- & -- & -- & \cellcolor{oursrowdark}70\% & \cellcolor{oursrowdark}60\% & 40\% & \cellcolor{oursrowdark}90\% & \cellcolor{oursrowdark}60\% & 50\%  \\
\midrule[0.6pt]
\multicolumn{11}{c}{\textit{\textbf{Multivariate Time Series Classification}}} \\
\midrule[0.6pt]
ArticularyWordRecognition & Acc. & 0.847 & -- & 0.948 & 0.897$^{\star}$ & -- & 0.944 & 0.853$^{\star}$ & -- & \textbf{0.949$^{\star}$}  \\
\rowcolor{gray!10} BasicMotions & Acc. & 0.775 & -- & 0.492 & 0.850$^{\star}$ & -- & 0.792$^{\star}$ & \textbf{0.900$^{\star}$} & -- & 0.758$^{\star}$  \\
CharacterTrajectories & Acc. & 0.957 & -- & 0.912 & 0.934 & -- & 0.902 & \textbf{0.961$^{\star}$} & -- & 0.915$^{\star}$  \\
\rowcolor{gray!10} FingerMovements & Acc. & 0.463 & -- & 0.500 & 0.480$^{\star}$ & -- & \textbf{0.507$^{\star}$} & 0.463 & -- & 0.503$^{\star}$  \\
SelfRegulationSCP2$\ddagger$ & Acc. & 0.507 & -- & 0.494 & 0.528$^{\star}$ & -- & 0.511$^{\star}$ & \textbf{0.535$^{\star}$} & -- & 0.489  \\
\rowcolor{gray!10} SpokenArabicDigits$\ddagger$ & Acc. & 0.946 & -- & 0.946 & 0.898 & -- & 0.943 & 0.938 & -- & \textbf{0.954$^{\star}$}  \\
\midrule[0.3pt]
\textit{Win rate vs.\ Base} & Acc. & -- & -- & -- & \cellcolor{oursrowdark}67\% & -- & 50\% & \cellcolor{oursrowdark}67\% & -- & \cellcolor{oursrowdark}83\%  \\
\bottomrule[1.3pt]
\end{tabular}
\end{adjustbox}
\end{table*}

\subsubsection{Evaluation and Implementation Details}

For evaluation, we use accuracy (\textbf{Acc.}) and F1-score (\textbf{F1}) . for classification, and \textbf{MSE} and \textbf{MAE} on normalized targets for forecasting.

The key hyperparameter settings are as follows. For forecasting, the input length and prediction horizon are set to 320 and 48 time steps, respectively. For latent action modeling, we set $d_{\mathrm{fuse}}=128$ (Sec.~\ref{subsubsec:fusion}), $d_\tau=16$ (Sec.~\ref{subsubsec:task_embed}), and $\beta=0.05$ (Eq.~\ref{eq:loss_action}). The VQ codebook (Sec.~\ref{sec:vq}) contains $K=16$ code vectors of dimension $d_{\mathrm{code}}=16$, with $M=2$ code selections per action. For the JEPA-based encoder (Sec.~\ref{sec:vjepa}), we set $d_{\mathrm{jepa}}=128$.

For the main experiments, we present representative results using selected downstream heads, averaged over three runs.
All experiments were conducted on two NVIDIA GB10 Grace Blackwell Superchips. Detailed settings are provided in Appendix~\ref{appendix:data_settings}, and full results are provided in Appendix~\ref{appendix:full}.

\begin{table}[!tb]
\centering
\footnotesize
\caption{Ablation results of ACTS-WM components using JEPA$_{\text{ACTS}}$ with the Obs.+Pred.\ variant, averaged over two selective downstream heads per dataset. The best results are highlighted in \textbf{bold}.}
\vspace{-1em}
\label{tb:ablation}
\setlength{\tabcolsep}{3pt}
\renewcommand{\arraystretch}{0.9}
\begin{adjustbox}{max width=\columnwidth, max totalheight=0.9\textheight}
\begin{tabular}{l*{8}{c}}
\toprule
\textit{Removed component} & \multicolumn{8}{c}{\textbf{ACTS-WM} (JEPA$_{\text{ACTS}}$, Obs.+Pred.)} \\
\cmidrule(lr){2-9}
Task embedding &  & \cmark &  &  &  & \cmark & \cmark & \cmark \\
Channel fusion &  &  & \cmark &  & \cmark &  & \cmark & \cmark \\
Temporal fusion &  &  &  & \cmark & \cmark & \cmark &  & \cmark \\
\cmidrule(lr){2-9}
\textbf{Dataset} & Full & (1) & (2) & (3) & (4) & (5) & (6) & (7) \\
\midrule[0.3pt]
\multicolumn{9}{c}{\textit{\textbf{Univariate Forecasting (MSE)}}} \\
\midrule[0.6pt]
Weather & \textbf{0.006} & \textbf{0.006} & \textbf{0.006} & \textbf{0.006} & \textbf{0.006} & \textbf{0.006} & \textbf{0.006} & \textbf{0.006} \\
\rowcolor{gray!10} ETTh2 & \textbf{0.452} & 0.479 & 0.484 & 0.478 & 0.468 & 0.477 & 0.487 & 0.478 \\
ETTm1 & 0.050 & \textbf{0.048} & 0.051 & 0.049 & 0.052 & 0.051 & 0.050 & 0.054 \\
\rowcolor{gray!10} ETTm2 & 0.188 & 0.193 & 0.184 & 0.183 & \textbf{0.180} & 0.189 & 0.192 & 0.182 \\
\midrule[0.3pt]
\midrule[0.3pt]
\textit{Average} & \textbf{0.174} & 0.181 & 0.181 & 0.179 & 0.176 & 0.181 & 0.184 & 0.180 \\
\midrule[0.6pt]
\multicolumn{9}{c}{\textit{\textbf{Multivariate Forecasting (MSE)}}} \\
\midrule[0.6pt]
ETTh1 & 0.905 & 1.103 & \textbf{0.901} & 0.987 & 0.938 & 1.102 & 1.101 & 1.098 \\
\rowcolor{gray!10} ETTh2 & \textbf{0.256} & 0.690 & 0.286 & 0.323 & 0.271 & 0.691 & 0.692 & 0.689 \\
ETTm1 & 0.576 & 0.787 & \textbf{0.564} & 0.689 & 0.599 & 0.789 & 0.800 & 0.799 \\
\rowcolor{gray!10} ETTm2 & \textbf{0.150} & 0.572 & 0.152 & 0.200 & 0.168 & 0.571 & 0.575 & 0.570 \\
\midrule[0.6pt]
\midrule[0.3pt]
\textit{Average} & \textbf{0.472} & 0.788 & 0.476 & 0.550 & 0.494 & 0.788 & 0.792 & 0.789 \\
\midrule[0.6pt]
\multicolumn{9}{c}{\textit{\textbf{Univariate Classification (Acc.)}}} \\
\midrule[0.3pt]
Earthquakes & 0.737 & 0.723 & 0.741 & \textbf{0.745} & 0.665 & 0.741 & 0.741 & 0.719 \\
\rowcolor{gray!10} UWaveGestureLibraryX & 0.645 & 0.642 & \textbf{0.663} & 0.645 & 0.643 & 0.636 & 0.636 & 0.655 \\
ShapesAll & 0.525 & 0.515 & 0.518 & 0.498 & 0.521 & \textbf{0.526} & 0.492 & 0.486 \\
\rowcolor{gray!10} EOGHorizontalSignal & 0.360 & 0.365 & 0.370 & 0.376 & \textbf{0.394} & 0.359 & 0.369 & 0.345 \\
\midrule[0.3pt]
\midrule[0.3pt]
\textit{Average} & 0.567 & 0.561 & \textbf{0.573} & 0.566 & 0.556 & 0.566 & 0.559 & 0.551 \\
\midrule[0.6pt]
\multicolumn{9}{c}{\textit{\textbf{Multivariate Classification (Acc.)}}} \\
\midrule[0.3pt]
ArticularyWordRecognition & \textbf{0.920} & 0.915 & 0.912 & 0.912 & 0.903 & 0.900 & 0.895 & 0.905 \\
\rowcolor{gray!10} BasicMotions & \textbf{0.788} & 0.725 & 0.712 & 0.662 & 0.663 & 0.713 & 0.738 & 0.712 \\
CharacterTrajectories & 0.920 & 0.932 & 0.930 & 0.918 & 0.930 & 0.917 & 0.920 & \textbf{0.936} \\
\rowcolor{gray!10} SpokenArabicDigits& 0.929 & 0.934 & 0.929 & 0.941 & \textbf{0.942} & 0.931 & 0.926 & 0.935 \\
\midrule[0.3pt]
\textit{Average} & \textbf{0.889} & 0.877 & 0.871 & 0.858 & 0.859 & 0.865 & 0.870 & 0.872 \\
\bottomrule[1.3pt]
\end{tabular}
\end{adjustbox}
\end{table}

\subsection{Performance Comparison}

Tables~\ref{tb:jepa_forecast} and~\ref{tb:jepa_classify} represent the results. Key observations are as follows:

\begin{itemize}[leftmargin=*]
    \item For the forecasting tasks in Table~\ref{tb:jepa_forecast}, ACTS-WM consistently improves forecasting performance compared to the base model of JEPA-based encoders (``Base''), particularly when using predicted target representations alone (``Pred. Only''). With our JEPA$_{\text{ACTS}}$, the ``Pred. Only'' variant reduces MSE on all 11 forecasting settings and achieves the lowest MSE across all five multivariate forecasting datasets. Similar improvements are observed with TS-JEPA and SC-JEPA, although their effectiveness varies across datasets and representation settings. Notably, on WikiTraffic, which is excluded from ACTS-WM pretraining, ``Pred. Only'' improves both MSE and MAE for all three JEPA-based encoders. Overall, ``Pred. Only'' generally outperforms ``Obs.+Pred.'' in forecasting, suggesting that the predicted target representations can improve time-series prediction.

    \item For classification, combining observed and predicted representations generally yields stronger results. With our JEPA$_{\text{ACTS}}$, ``Obs.+Pred.'' improves accuracy over Base on 9 out of 10 univariate classification datasets, all of which are excluded from ACTS-WM pretraining. For multivariate classification, JEPA$_{\text{ACTS}}$ with ``Obs.+Pred.'' achieves the best accuracy on 3 out of 6 datasets, while SC-JEPA with ``Obs.+Pred.'' improves upon its Base on 5 out of 6 datasets. These results suggest that predicted representations of ACTS-WM can complement observed representations for classification. Furthermore, the improvements on datasets excluded from ACTS-WM pretraining indicate the potential of the learned world model to generalize across datasets.
\end{itemize}

\suppressfloats[t]
\begin{figure}[t]
\graphicspath{{figs/}}
\centering
\includegraphics[width=0.98\columnwidth]{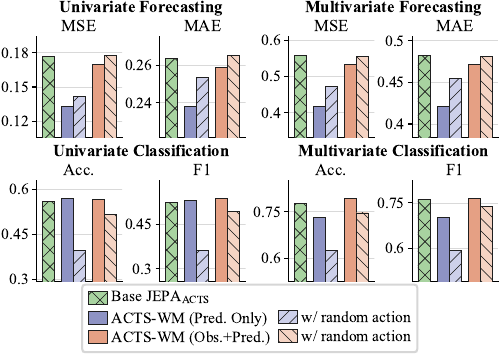}
\vspace{-0.7em}
\caption{Comparison of learned and randomly assigned latent actions in ACTS-WM using JEPA$_{\text{ACTS}}$.}
\label{fig:random_action}
\end{figure}

\begin{figure}[!b]
\graphicspath{{figs/}}
\centering
\subfloat[Training loss of the world model ($\mathcal{L}_{\mathrm{ACTS-WM}}$) for codebook sizes $K\in\{2,4,16,64,128\}$ with with JEPA$_{\text{ACTS}}$.\label{fig:wm_loss}]{%
  \includegraphics[width=0.98\columnwidth]{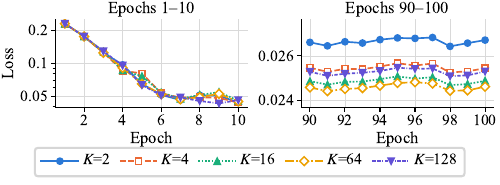}}
\\[2pt]
\subfloat[Forecasting MSE for different prediction horizons $H$.\label{fig:horizon-bar}]{%
  \includegraphics[width=\columnwidth]{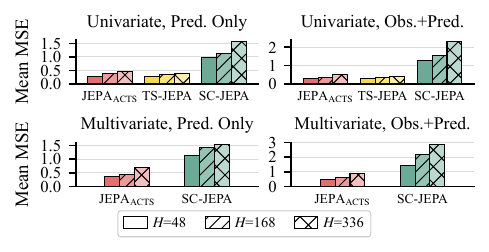}}
\vspace{-0.8em}
\caption{Parameter analysis for the codebook size and prediction horizons.}
\label{fig:param_analysis}
\end{figure}

\subsection{Ablation Study}

We conduct ablation studies to evaluate the proposed designs in Secs.~\ref{subsubsec:fusion} and~\ref{subsubsec:task_embed}, and latent action selection in Sec.~\ref{sec:vq}.

\noindent\textbf{Component Ablation.}
Table~\ref{tb:ablation} evaluates the contributions of designs in Secs.~\ref{subsubsec:fusion} and~\ref{subsubsec:task_embed}. For the average performance, the full ACTS-WM achieves the best in three of the four task categories. Task embeddings are particularly important for multivariate forecasting, as removing them increases the average MSE from 0.472 to 0.788. Temporal fusion also contributes to multivariate tasks, with its removal increasing forecasting MSE from 0.472 to 0.550 and reducing classification accuracy from 0.889 to 0.858. In contrast, removing channel fusion has a small impact on multivariate forecasting and slightly improves univariate classification accuracy. Overall, the effectiveness of each component varies across different tasks and datasets.

\begin{figure*}[t]
\graphicspath{{figs/}}
\centering
\subfloat[Forecasting datasets. Colors indicate quartiles of relative mean differences between observed and target time series, ordered from smallest to largest.]{%
\includegraphics[width=0.9\textwidth]{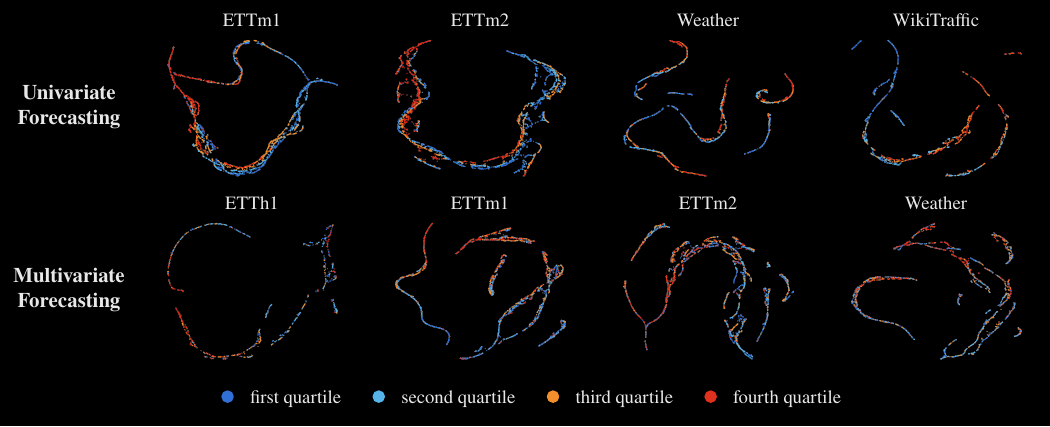}
\label{fig:umap_forecasting}}
\\[2pt]
\subfloat[Classification datasets. Different colors represent different class labels within each dataset.]{%
\includegraphics[width=0.9\textwidth]{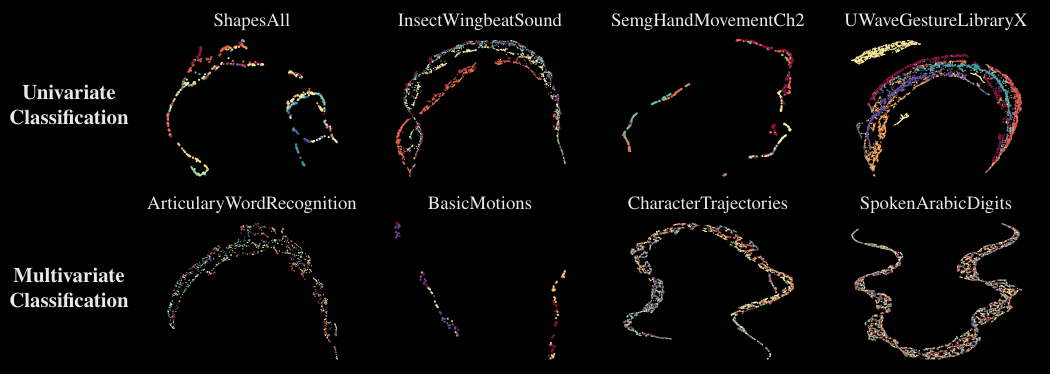}
\label{fig:umap_classification}}
\vspace{-1em}
\caption{UMAP projection of continuous latent actions inferred by the IDM before quantization. Each point represents a latent action inferred from an observation pair.}
\label{fig:umap}
\end{figure*}

\noindent\textbf{Effect of Learned Latent Actions.}
To evaluate latent action selection (Sec.~\ref{sec:vq}), we compare ACTS-WM using learned latent actions with randomly assigned actions in Figure~\ref{fig:random_action}. Replacing learned actions with random actions consistently degrades performance across all four task categories.
For forecasting, ``Pred. Only'' achieves the lowest errors, while random actions increase both MSE and MAE. For classification, ``Pred. Only'' is sensitive to random actions, with substantial decreases in accuracy and F1-score. Nevertheless, using random actions generally degrades performance, often below the Base JEPA$_{\text{ACTS}}$ encoder, particularly in classification tasks. These findings suggest that learned latent actions contribute meaningful information and that combining observed and predicted representations can improve robustness.

\subsection{Parameter Analysis}

\noindent\textbf{Effect of Codebook Size.}
We investigate the effect of codebook size on world model training over 100 epochs. In Figure~\ref{fig:wm_loss}, the training loss decreases rapidly during the first 10 epochs across all settings. However, differences emerge toward the end of training. Increasing the codebook size from 2 to 64 generally reduces the final training loss, with the lowest loss achieved at 64. Further increasing the size to 128 slightly increases the loss. These results suggest that an action space that is too small ($K=2$) may fail to capture sufficient transition information, while an excessively large action space ($K=128$) may also hinder representation prediction.

\noindent\textbf{Prediction Horizon.}
Figure~\ref{fig:horizon-bar} compares forecasting performance across prediction horizons of $H\in\{48,168,336\}$. The mean MSE increases with longer horizons across all JEPA-based encoders in both univariate and multivariate settings. This degradation is particularly pronounced for SC-JEPA, especially under the ``Obs.+Pred.'' setting. Overall, these results highlight the increasing difficulty of modeling longer-term time-series transitions.

\subsection{Interpretation of Learned Latent Actions}

To investigate what the learned latent actions capture, we visualize the continuous latent representations inferred by the IDM before quantization (Eq.~\ref{eq:idm}) using UMAP projection~\cite{mcinnes2018umap}, as shown in Figure~\ref{fig:umap}. We select representative forecasting and classification datasets with sufficient evaluation data samples.

For forecasting (Figure~\ref{fig:umap}(a)), the first (smallest difference) and fourth quartiles (largest difference) exhibit distinct distributions across several datasets, suggesting that the learned latent actions capture information related to the magnitude of changes between observed and future time series. For classification, Figure~\ref{fig:umap}(b) reveals class-dependent structures in several datasets, although some classes exhibit overlapping distributions. These observations suggest that the IDM captures transition characteristics associated with different time-series classes. 

Overall, the visualizations provide qualitative evidence that the learned latent actions encode meaningful information about time-series transitions across forecasting and classification tasks.
\section{Conclusion and Future Work}

In this work, we proposed ACTS-WM, a self-supervised latent action-conditioned world modeling framework for time-series analysis. ACTS-WM learns discrete latent actions used to condition a JEPA-based world model for predicting target representations. Experiments on 22 datasets covering univariate and multivariate forecasting and classification show improvements over pretrained JEPA-based encoders on many datasets. Further analyses highlight the benefits of learned latent actions and action conditioning. Future work will focus on improving the latent policy's ability to predict latent actions.

\bibliographystyle{ACM-Reference-Format}
\bibliography{reference}

@String{Computing = "Computing" }

@String{Chelsea = "Chelsea" }

@String{Springer = "Springer-Verlag" }

@inproceedings{schmidt2024learning,
  title={Learning to act without actions},
  author={Schmidt, Dominik and Jiang, Minqi},
  booktitle={International Conference on Learning Representations},
  volume={2024},
  pages={9379--9395},
  year={2024}
}

@inproceedings{ye2025latent,
  title={Latent action pretraining from videos},
  author={Ye, Seonghyeon and Jang, Joel and Jeon, Byeongguk and Joo, Se June and Yang, Jianwei and Peng, Baolin and Mandlekar, Ajay and Tan, Reuben and Chao, Yu-Wei and Lin, Bill Yuchen and others},
  booktitle={International Conference on Learning Representations},
  volume={2025},
  pages={28213--28239},
  year={2025}
}

@inproceedings{
gao2025adaworld,
title={AdaWorld: Learning Adaptable World Models with Latent Actions},
author={Shenyuan Gao and Siyuan Zhou and Yilun Du and Jun Zhang and Chuang Gan},
booktitle={Forty-second International Conference on Machine Learning},
year={2025},
url={https://openreview.net/forum?id=QQegZj99sk}
}

@inproceedings{fraikin2024t,
  title={T-rep: Representation learning for time series using time-embeddings},
  author={Fraikin, Archibald and Bennetot, Adrien and Allassonni{\`e}re, St{\'e}phanie},
  booktitle={International Conference on Learning Representations},
  volume={2024},
  pages={31285--31305},
  year={2024}
}

@inproceedings{yue2022ts2vec,
  title={Ts2vec: Towards universal representation of time series},
  author={Yue, Zhihan and Wang, Yujing and Duan, Juanyong and Yang, Tianmeng and Huang, Congrui and Tong, Yunhai and Xu, Bixiong},
  booktitle={Proceedings of the AAAI conference on artificial intelligence},
  volume={36},
  number={8},
  pages={8980--8987},
  year={2022}
}

@inproceedings{goswami2024moment,
    author = {Goswami, Mononito and Szafer, Konrad and Choudhry, Arjun and Cai, Yifu and Li, Shuo and Dubrawski, Artur},
    title = {MOMENT: a family of open time-series foundation models},
    year = {2024},
    publisher = {JMLR.org},
    booktitle = {Proceedings of the 41st International Conference on Machine Learning},
    articleno = {642},
    numpages = {38}
}

@inproceedings{nie2023a,
    title={A Time Series is Worth 64 Words:  Long-term Forecasting with Transformers},
    author={Yuqi Nie and Nam H Nguyen and Phanwadee Sinthong and Jayant Kalagnanam},
    booktitle={The Eleventh International Conference on Learning Representations },
    year={2023},
    url={https://openreview.net/forum?id=Jbdc0vTOcol}
}

@InProceedings{pmlr-v235-woo24a,
  title = 	 {Unified Training of Universal Time Series Forecasting Transformers},
  author =       {Woo, Gerald and Liu, Chenghao and Kumar, Akshat and Xiong, Caiming and Savarese, Silvio and Sahoo, Doyen},
  booktitle = 	 {Proceedings of the 41st International Conference on Machine Learning},
  pages = 	 {53140--53164},
  year = 	 {2024},
  volume = 	 {235},
  series = 	 {Proceedings of Machine Learning Research},
  month = 	 {21--27 Jul},
  publisher =    {PMLR}
}

@article{assran2025v,
  title={V-jepa 2: Self-supervised video models enable understanding, prediction and planning},
  author={Assran, Mido and Bardes, Adrien and Fan, David and Garrido, Quentin and Howes, Russell and Muckley, Matthew and Rizvi, Ammar and Roberts, Claire and Sinha, Koustuv and Zholus, Artem and others},
  journal={arXiv preprint arXiv:2506.09985},
  year={2025}
}

@inproceedings{
    ennadir2024joint,
    title={Joint Embedding go Temporal},
    author={Sofiane ENNADIR and Siavash Golkar and Leopoldo Sarra},
    booktitle={NeurIPS Workshop on Time Series in the Age of Large Models},
    year={2024},
    url={https://openreview.net/forum?id=FIdbozebmy}
}

@inproceedings{
    he2026scjepa,
    title={{SC}-{JEPA}: Stabilizing Latent Predictive Learning for Time-Series Anomaly Prediction},
    author={Yanan He and Yunshi Wen and Xin Wang and Tengfei Ma},
    booktitle={Forecasting as a New Frontier of Intelligence},
    year={2026},
    url={https://openreview.net/forum?id=SNjhsnUvhH}
}

@inproceedings{
    dutta2026giving,
    title={Giving Sensors a Voice: Multimodal {JEPA} for Semantic Time-Series Embeddings},
    author={Utsav Dutta and Gerardo Pastrana and Sina Khoshfetrat Pakazad and Henrik Ohlsson},
    booktitle={Forty-third International Conference on Machine Learning},
    year={2026},
    url={https://openreview.net/forum?id=lNoaqrOXti}
}

@article{lecun2022path,
  title={A path towards autonomous machine intelligence version 0.9. 2, 2022-06-27},
  author={LeCun, Yann and others},
  journal={Open Review},
  volume={62},
  number={1},
  pages={1--62},
  year={2022}
}

@inproceedings{lin2021temporal,
  title={Temporal convolutional attention neural networks for time series forecasting},
  author={Lin, Yang and Koprinska, Irena and Rana, Mashud},
  booktitle={2021 International joint conference on neural networks (IJCNN)},
  pages={1--8},
  year={2021},
  organization={IEEE}
}

@inproceedings{khorrambakht2026worldplanner,
  title={Worldplanner: Monte carlo tree search and mpc with action-conditioned visual world models},
  author={Ortiz-Haro, Joaquim and Amigo, Joseph and Mostafa, Omar and Dugas, Daniel and Meier, Franziska and Righetti, Ludovic},
  booktitle={2026 IEEE International Conference on Robotics and Automation (ICRA)},
  pages={22166--22173},
  year={2026},
  organization={IEEE}
}

@inproceedings{wu2023daydreamer,
  title={Daydreamer: World models for physical robot learning},
  author={Wu, Philipp and Escontrela, Alejandro and Hafner, Danijar and Abbeel, Pieter and Goldberg, Ken},
  booktitle={Conference on robot learning},
  pages={2226--2240},
  year={2023},
  organization={PMLR}
}

@inproceedings{guo2026ctrl,
  title={Ctrl-world: A controllable generative world model for robot manipulation},
  author={Guo, Yanjiang and Shi, Lucy and Chen, Jianyu and Finn, Chelsea},
  booktitle={International Conference on Learning Representations},
  volume={2026},
  pages={6121--6138},
  year={2026}
}

@article{bai2018empirical,
  title={An empirical evaluation of generic convolutional and recurrent networks for sequence modeling},
  author={Bai, Shaojie and Kolter, J Zico and Koltun, Vladlen},
  journal={arXiv preprint arXiv:1803.01271},
  year={2018}
}

@inproceedings{ronneberger2015u,
  title={U-net: Convolutional networks for biomedical image segmentation},
  author={Ronneberger, Olaf and Fischer, Philipp and Brox, Thomas},
  booktitle={International Conference on Medical image computing and computer-assisted intervention},
  pages={234--241},
  year={2015},
  organization={Springer}
}

@article{mcinnes2018umap,
  title={Umap: Uniform manifold approximation and projection for dimension reduction},
  author={McInnes, Leland and Healy, John and Melville, James},
  journal={arXiv preprint arXiv:1802.03426},
  year={2018}
}

@inproceedings{
petersen2026hepa,
title={{HEPA}: A Self-Supervised Horizon-Conditioned Event Predictive Architecture for Time Series},
author={Jonas Petersen and Gian-Alessandro Lombardi and Riccardo Maggioni and Camilla Mazzoleni and Federico Martelli and Philipp Christian Petersen},
booktitle={2nd ICML Workshop on Foundation Models for Structured Data},
year={2026},
url={https://openreview.net/forum?id=TdpqVBB8na}
}

@misc{qi2026staticforecastingunleashingpower,
      title={Beyond Static Forecasting: Unleashing the Power of World Models for Mobile Traffic Extrapolation}, 
      author={Xiaoqian Qi and Haoye Chai and Yue Wang and Yong Li},
      year={2026},
      eprint={2604.08199},
      archivePrefix={arXiv},
      primaryClass={cs.NI},
      url={https://arxiv.org/abs/2604.08199}, 
}

@inproceedings{
shaj2023multi,
title={Multi Time Scale World Models},
author={Vaisakh Shaj and Saleh GHOLAM ZADEH and Ozan Demir and Luiz Ricardo Douat and Gerhard Neumann},
booktitle={Thirty-seventh Conference on Neural Information Processing Systems},
year={2023},
url={https://openreview.net/forum?id=fY7dShbtmo}
}

@article{lu2026policy,
  title={Policy and World Modeling Co-Training for Language Agents},
  author={Lu, Ning and Lin, Baijiong and Liu, Shengcai and Wu, Jiahao and Lv, Haoze and Wei, Yanbin and Zhu, Lingting and Qian, Shengju and Wang, Xin and Chen, Ying-Cong and others},
  journal={arXiv preprint arXiv:2606.02388},
  year={2026}
}

@inproceedings{
    wang2025co,
    title={Co-Evolving Latent Action World Models},
    author={Yucen Wang and Fengming Zhang and De-Chuan Zhan and Li Zhao and Kaixin Wang and Jiang Bian},
    booktitle={Forty-third International Conference on Machine Learning},
    year={2026},
    url={https://openreview.net/forum?id=iec4FIZyrx}
}

@article{bagnall2018uea,
  title={The UEA multivariate time series classification archive, 2018},
  author={Bagnall, Anthony and Dau, Hoang Anh and Lines, Jason and Flynn, Michael and Large, James and Bostrom, Aaron and Southam, Paul and Keogh, Eamonn},
  journal={arXiv preprint arXiv:1811.00075},
  year={2018}
}

@inproceedings{NEURIPS2021_bcc0d400,
 author = {Wu, Haixu and Xu, Jiehui and Wang, Jianmin and Long, Mingsheng},
 booktitle = {Advances in Neural Information Processing Systems},
 pages = {22419--22430},
 publisher = {Curran Associates, Inc.},
 title = {Autoformer: Decomposition Transformers with Auto-Correlation for Long-Term Series Forecasting},
 volume = {34},
 year = {2021}
}

@ARTICLE{8894743,
  author={Dau, Hoang Anh and Bagnall, Anthony and Kamgar, Kaveh and Yeh, Chin-Chia Michael and Zhu, Yan and Gharghabi, Shaghayegh and Ratanamahatana, Chotirat Ann and Keogh, Eamonn},
  journal={IEEE/CAA Journal of Automatica Sinica}, 
  title={The UCR time series archive}, 
  year={2019},
  volume={6},
  number={6},
  pages={1293-1305},
  doi={10.1109/JAS.2019.1911747}
}

@inproceedings{ICLR2024_fb9d01fb,
 author = {Early, Joseph and Cheung, Gavin and Cutajar, Kurt and Xie, Hanting and Kandola, Jas and Twomey, Niall},
 booktitle = {International Conference on Learning Representations},
 pages = {56800--56828},
 title = {Inherently Interpretable Time Series Classification via Multiple Instance Learning},
 volume = {2024},
 year = {2024}
}

@inproceedings{godahewa2021monash,
title={Monash Time Series Forecasting Archive},
author={Rakshitha Wathsadini Godahewa and Christoph Bergmeir and Geoffrey I. Webb and Rob Hyndman and Pablo Montero-Manso},
booktitle={Thirty-fifth Conference on Neural Information Processing Systems Datasets and Benchmarks Track (Round 2)},
year={2021},
url={https://openreview.net/forum?id=wEc1mgAjU-}
}

@misc{
cohen2024toto,
title={Toto: Time Series Optimized Transformer for Observability},
author={Ben Cohen and Emaad Khwaja and Kan Wang and Charles Masson and Elise Ram{\'e} and Youssef Doubli and Othmane Abou-Amal},
year={2024},
url={https://openreview.net/forum?id=ayupWYA1qD}
}

@article{liu2025moirai,
  title={Moirai 2.0: When less is more for time series forecasting},
  author={Liu, Chenghao and Aksu, Taha and Liu, Juncheng and Liu, Xu and Yan, Hanshu and Pham, Quang and Savarese, Silvio and Sahoo, Doyen and Xiong, Caiming and Li, Junnan},
  journal={arXiv preprint arXiv:2511.11698},
  year={2025}
}

@inproceedings{
liu2025moiraimoe,
title={Moirai-MoE: Empowering Time Series Foundation Models with Sparse Mixture of Experts},
author={Xu Liu and Juncheng Liu and Gerald Woo and Taha Aksu and Yuxuan Liang and Roger Zimmermann and Chenghao Liu and Junnan Li and Silvio Savarese and Caiming Xiong and Doyen Sahoo},
booktitle={Forty-second International Conference on Machine Learning},
year={2025},
url={https://openreview.net/forum?id=SrEOUSyJcR}
}

@inproceedings{NEURIPS2024_fe248e22,
 author = {Gao, Shanghua and Koker, Teddy and Queen, Owen and Hartvigsen, Thomas and Tsiligkaridis, Theodoros and Zitnik, Marinka},
 booktitle = {Advances in Neural Information Processing Systems},
 doi = {10.52202/079017-4463},
 pages = {140589--140631},
 publisher = {Curran Associates, Inc.},
 title = {UniTS: A Unified Multi-Task Time Series Model},
 volume = {37},
 year = {2024}
}

@inproceedings{
shi2025timemoe,
title={Time-MoE: Billion-Scale Time Series Foundation Models with Mixture of Experts},
author={Xiaoming Shi and Shiyu Wang and Yuqi Nie and Dianqi Li and Zhou Ye and Qingsong Wen and Ming Jin},
booktitle={The Thirteenth International Conference on Learning Representations},
year={2025},
url={https://openreview.net/forum?id=e1wDDFmlVu}
}

@misc{hafner2025trainingagentsinsidescalable,
      title={Training Agents Inside of Scalable World Models}, 
      author={Danijar Hafner and Wilson Yan and Timothy Lillicrap},
      year={2025},
      eprint={2509.24527},
      archivePrefix={arXiv},
      primaryClass={cs.AI},
      url={https://arxiv.org/abs/2509.24527}, 
}

@inproceedings{NEURIPS2024_6bdde037,
 author = {Alonso, Eloi and Jelley, Adam and Micheli, Vincent and Kanervisto, Anssi and Storkey, Amos and Pearce, Tim and Fleuret, Fran\c{c}ois},
 booktitle = {Advances in Neural Information Processing Systems},
 doi = {10.52202/079017-1873},
 pages = {58757--58791},
 publisher = {Curran Associates, Inc.},
 title = {Diffusion for World Modeling: Visual Details Matter in Atari},
 volume = {37},
 year = {2024}
}

@article{agarwal2026cosmos,
  title={Cosmos 3: Omnimodal world models for physical ai},
  author={Agarwal, Niket and Ali, Arslan and Allen, Jon and Antolini, Martin and Aubame, Adeline and Azzolini, Alisson and Bai, Junjie and Bala, Maciej and Balaji, Yogesh and Bapst, Josh and others},
  journal={arXiv preprint arXiv:2606.02800},
  year={2026}
}

@article{dai2026pythia,
  title={Pythia: Toward Foundation World Models for Multimodal Time Series},
  author={Dai, Xilin and Chen, Hongzhou and Hu, Yifan and Liu, Yiding and Dong, Zewei and Yang, Jiang-Ming},
  journal={arXiv preprint arXiv:2610.05240},
  year={2026}
}

@misc{verdenius2024latpfnjointembeddingpredictive,
      title={LaT-PFN: A Joint Embedding Predictive Architecture for In-context Time-series Forecasting}, 
      author={Stijn Verdenius and Andrea Zerio and Roy L. M. Wang},
      year={2024},
      eprint={2405.10093},
      archivePrefix={arXiv},
      primaryClass={cs.LG},
      url={https://arxiv.org/abs/2405.10093}, 
}

@article{zhang2026h,
  title={H-JEPA: End-to-end learning of hierarchical world models for visual planning},
  author={Zhang, Wancong and Terver, Basile and Rabbat, Michael and LeCun, Yann and Balestriero, Randall},
  journal={arXiv preprint arXiv:2610.06805},
  year={2026}
}

@article{gray1984vector,
  title={Vector quantization},
  author={Gray, Robert},
  journal={IEEE Assp Magazine},
  volume={1},
  number={2},
  pages={4--29},
  year={1984},
  publisher={IEEE}
}

@dataset{godahewa_2022_7371038,
  author       = {Godahewa, Rakshitha and
                  Bergmeir, Christoph and
                  Webb, Geoff and
                  Hyndman, Rob and
                  Montero-Manso, Pablo},
  title        = {Extended Wikipedia Web Traffic Daily Dataset
                   (without Missing Values)
                  },
  month        = nov,
  year         = 2022,
  publisher    = {Zenodo},
  version      = 1,
  doi          = {10.5281/zenodo.7371038},
  url          = {https://doi.org/10.5281/zenodo.7371038},
}

@inproceedings{
bruce2024genie,
title={Genie: Generative Interactive Environments},
author={Jake Bruce and Michael D Dennis and Ashley Edwards and Jack Parker-Holder and Yuge Shi and Edward Hughes and Matthew Lai and Aditi Mavalankar and Richie Steigerwald and Chris Apps and Yusuf Aytar and Sarah Maria Elisabeth Bechtle and Feryal Behbahani and Stephanie C.Y. Chan and Nicolas Heess and Lucy Gonzalez and Simon Osindero and Sherjil Ozair and Scott Reed and Jingwei Zhang and Konrad Zolna and Jeff Clune and Nando de Freitas and Satinder Singh and Tim Rockt{\"a}schel},
booktitle={Forty-first International Conference on Machine Learning},
year={2024},
url={https://openreview.net/forum?id=bJbSbJskOS}
}

@inproceedings{
fu2025on,
title={On the Integration of Spatial-Temporal Knowledge: A Lightweight Approach to Atmospheric Time Series Forecasting},
author={Yisong Fu and Fei Wang and Zezhi Shao and Boyu Diao and Lin Wu and Zhulin An and Chengqing Yu and Yujie Li and Yongjun Xu},
booktitle={The Thirty-ninth Annual Conference on Neural Information Processing Systems},
year={2025},
url={https://openreview.net/forum?id=thHhKPlt8q}
}

@inproceedings{zhou2021informer,
  title={Informer: Beyond efficient transformer for long sequence time-series forecasting},
  author={Zhou, Haoyi and Zhang, Shanghang and Peng, Jieqi and Zhang, Shuai and Li, Jianxin and Xiong, Hui and Zhang, Wancai},
  booktitle={Proceedings of the AAAI conference on artificial intelligence},
  volume={35},
  number={12},
  pages={11106--11115},
  year={2021}
}

@INPROCEEDINGS{9414560,
  author={Shi, Yangyang and Wang, Yongqiang and Wu, Chunyang and Yeh, Ching-Feng and Chan, Julian and Zhang, Frank and Le, Duc and Seltzer, Mike},
  booktitle={ICASSP 2021 - 2021 IEEE International Conference on Acoustics, Speech and Signal Processing (ICASSP)}, 
  title={Emformer: Efficient Memory Transformer Based Acoustic Model for Low Latency Streaming Speech Recognition}, 
  year={2021},
  volume={},
  number={},
  pages={6783-6787},
  doi={10.1109/ICASSP39728.2021.9414560}}

@inproceedings{10.1145/3774904.3792374,
author = {Zhang, Jianqi and Wang, Jingyao and Qiang, Wenwen and Xu, Fanjiang and Zheng, Changwen},
title = {Enhancing Large Language Models for Time-Series Forecasting via Vector-Injected In-Context Learning},
year = {2026},
isbn = {9798400723070},
publisher = {Association for Computing Machinery},
address = {New York, NY, USA},
url = {https://doi.org/10.1145/3774904.3792374},
doi = {10.1145/3774904.3792374},
booktitle = {Proceedings of the ACM Web Conference 2026},
pages = {7211–7222},
numpages = {12},
location = {Dubai, United Arab Emirates},
series = {WWW '26}
}

@inproceedings{10.1145/3292500.3330680,
author = {Ren, Hansheng and Xu, Bixiong and Wang, Yujing and Yi, Chao and Huang, Congrui and Kou, Xiaoyu and Xing, Tony and Yang, Mao and Tong, Jie and Zhang, Qi},
title = {Time-Series Anomaly Detection Service at Microsoft},
year = {2019},
isbn = {9781450362016},
publisher = {Association for Computing Machinery},
address = {New York, NY, USA},
url = {https://doi.org/10.1145/3292500.3330680},
doi = {10.1145/3292500.3330680},
booktitle = {Proceedings of the 25th ACM SIGKDD International Conference on Knowledge Discovery \& Data Mining},
pages = {3009–3017},
numpages = {9},
location = {Anchorage, AK, USA},
series = {KDD '19}
}

\appendix
\section{Appendix}

\subsection{Dataset for Pretraining and Evaluation}

All datasets are listed in Table~\ref{tb:datasets}.

\begin{table}[t]
\centering
\small
\caption{Datasets used for ACTS-WM pretraining and downstream evaluation. Univariate and multivariate classification datasets are from UCR~\cite{8894743} and UEA~\cite{bagnall2018uea}, respectively, except WebTraffic~\cite{ICLR2024_fb9d01fb}. Forecasting datasets are from Autoformer~\cite{NEURIPS2021_bcc0d400}, except WikiTraffic~\cite{godahewa2021monash,godahewa_2022_7371038}. (\#T refers to the number of time steps, and \#C refers to the number of channels).}
\label{tb:datasets}
\renewcommand{\arraystretch}{1.2}
\resizebox{\columnwidth}{!}{
\begin{tabular}{lccrr}
\toprule
\textbf{Dataset} & \textbf{Pretrain } & \textbf{Evaluation} & \textbf{\#T} & \textbf{\#C} \\
\midrule
\multicolumn{5}{l}{\textit{Univariate classification}$^{*}$} \\
Computers                   &        & \cmark & 180,000 & 1   \\
Earthquakes                 &        & \cmark & 164,864 & 1   \\
EOGHorizontalSignal         &        & \cmark & 452,500 & 1   \\
EthanolLevel                &        & \cmark & 882,504 & 1   \\
InsectWingbeatSound         &        & \cmark & 56,320  & 1   \\
SemgHandMovementCh2         &        & \cmark & 675,000 & 1   \\
ShapesAll                   &        & \cmark & 307,200 & 1   \\
UWaveGestureLibraryX        &        & \cmark & 282,240 & 1   \\
WormsTwoClass               &        & \cmark & 162,900 & 1   \\
\addlinespace[2pt]
WebTraffic                  &        & \cmark & 504,000 & 1   \\
\midrule
\multicolumn{5}{l}{\textit{Multivariate classification}$^{*}$} \\
SelfRegulationSCP2          &        & \cmark & 230,400 & 7   \\
SpokenArabicDigits          &        & \cmark & 263,224 & 13  \\
ArticularyWordRecognition       & \cmark & \cmark & 39,600  & 9   \\
BasicMotions                & \cmark & \cmark & 4,000   & 6   \\
CharacterTrajectories       & \cmark & \cmark & 169,218 & 3   \\
FingerMovements             & \cmark & \cmark & 15,800  & 28  \\
AtrialFibrillation          & \cmark &        & 9,600   & 2   \\
Cricket                     & \cmark &        & 129,276 & 6   \\
Epilepsy                    & \cmark &        & 28,222  & 3   \\
ERing                       & \cmark &        & 1,950   & 4   \\
HandMovementDirection       & \cmark &        & 64,000  & 10  \\
Handwriting                 & \cmark &        & 22,800  & 3   \\
Libras                      & \cmark &        & 8,100   & 2   \\
LSST                        & \cmark &        & 88,524  & 6   \\
NATOPS                      & \cmark &        & 9,180   & 24  \\
RacketSports                & \cmark &        & 4,530   & 6   \\
SelfRegulationSCP1          & \cmark &        & 240,128 & 6   \\
UWaveGestureLibrary         & \cmark &        & 37,800  & 3   \\
\midrule
\multicolumn{5}{l}{\textit{Forecasting}} \\
ETTh1                       & \cmark & \cmark & 17,420  & 7   \\
ETTh2                       & \cmark & \cmark & 17,420  & 7   \\
ETTm1                       & \cmark & \cmark & 69,680  & 7   \\
ETTm2                       & \cmark & \cmark & 69,680  & 7   \\
Weather                     & \cmark & \cmark & 52,696  & 21  \\
WikiTraffic$^{*}$           &        & \cmark & 56,254  & 1   \\
\bottomrule
\multicolumn{5}{l}{\footnotesize $^{*}$These Datasets are provided as windowed instances. The total number of time steps} \\
\multicolumn{5}{l}{\footnotesize \phantom{$^{*}$} is estimated by multiplying the average sequence length by the number of samples.} \\
\end{tabular}
}
\end{table}

\subsection{Detailed Data Preprocessing and Settings}
\label{appendix:data_settings}

\subsubsection{Preprocessing}

For the datasets and variant lengths in the forecasting tasks, we impute the last value to the maximum length among all samples.
Following~\cite{yue2022ts2vec}, for univariate forecasting tasks, the target value is set as \emph{oil temperature} for ETDatasets and \emph{co2} for the Weather dataset.

To prevent data leakage, we use only the training split during pretraining for datasets included in downstream evaluation. For other datasets, both training and test splits are used for pretraining. For forecasting tasks, datasets are divided into training, validation, and test sets at a ratio of 65:5:30. For classification tasks, we retain the original sequence lengths and predefined train/test splits, reserving 10\% of the training set for validation.

\subsubsection{Settings for JEPA-based Encoders}

For pretraining ACTS-WM with our JEPA$_{\text{ACTS}}$, we use 21 datasets, including 16 from the UEA archive~\cite{bagnall2018uea}, 4 ETT datasets, and the Weather dataset. The ETT and Weather datasets are widely used forecasting benchmarks available through the Autoformer repository~\cite{NEURIPS2021_bcc0d400}. 

Since TS-JEPA and SC-JEPA require fixed-length inputs, we pretrain ACTS-WM separately with each encoder. Specifically, we use 12 univariate datasets for TS-JEPA, which supports only univariate inputs, and 22 datasets for SC-JEPA.

For downstream evaluation, we use 22 datasets covering forecasting and classification. These include 9 UCR datasets~\cite{8894743} and WebTraffic~\cite{ICLR2024_fb9d01fb} for univariate classification, 6 UEA datasets for multivariate classification, and 6 forecasting datasets: ETTh1, ETTh2, ETTm1, ETTm2, Weather, and WikiTraffic~\cite{godahewa2021monash, godahewa_2022_7371038}.

\subsection{Model Settings}
\label{appendix:model_settings}

\subsubsection{Downstream Head Settings}

We adapt the JEPA encoders and our ACTS-WM to downstream tasks using six different downstream heads. We use three classification heads: linear head, SVM head, and MLP head; and three forecasting heads: linear head, ridge regression head, and NLH head.
For forecasting, linear head and NLH head follow the configurations described in CHARM~\cite{dutta2026giving}. The ridge head is a per-channel ridge regression, mapping the flattened tokens to the forecast horizon. It is solved in closed form, with the regularization strength selected based on validation performance.

\subsubsection{Hyperparameter Settings}

The key hyperparameters of ACTS-WM are configured as follows. 

\begin{itemize}[leftmargin=*]

\item For latent action modeling, we set $d_{\mathrm{fuse}}=128$ (Sec.~\ref{subsubsec:fusion}), $d_\tau=16$ (Sec.~\ref{subsubsec:task_embed}), and $\beta=0.05$ (Eq.~\ref{eq:loss_action}). The VQ codebook (Sec.~\ref{sec:vq}) contains $K=16$ code vectors of dimension $d_{\mathrm{code}}=16$, with $M=2$ code selections per action. For the JEPA-based encoder (Sec.~\ref{sec:vjepa}), we set $d_{\mathrm{jepa}}=128$.

\item For the settings of JEPA-encoders, We train the JEPA$_{\text{ACTS-WM}}$ encoder for 200 epochs, with the learning rate initially set to $3 \times 10^{-5}$, linearly warmed up to $3 \times 10^{-4}$ over the first 5\% of training epochs, then gradually decaying to $1 \times 10^{-6}$ with a cosine learning rate schedule. For other JEPA-based time series encoders, we pretrain TS-JEPA~\cite{ennadir2024joint} and SC-JEPA\cite{he2026scjepa} following the configurations specified in their respective papers.

\end{itemize}

\begin{figure}[t]
\graphicspath{{figs/}}
\centering
\includegraphics[width=\columnwidth]{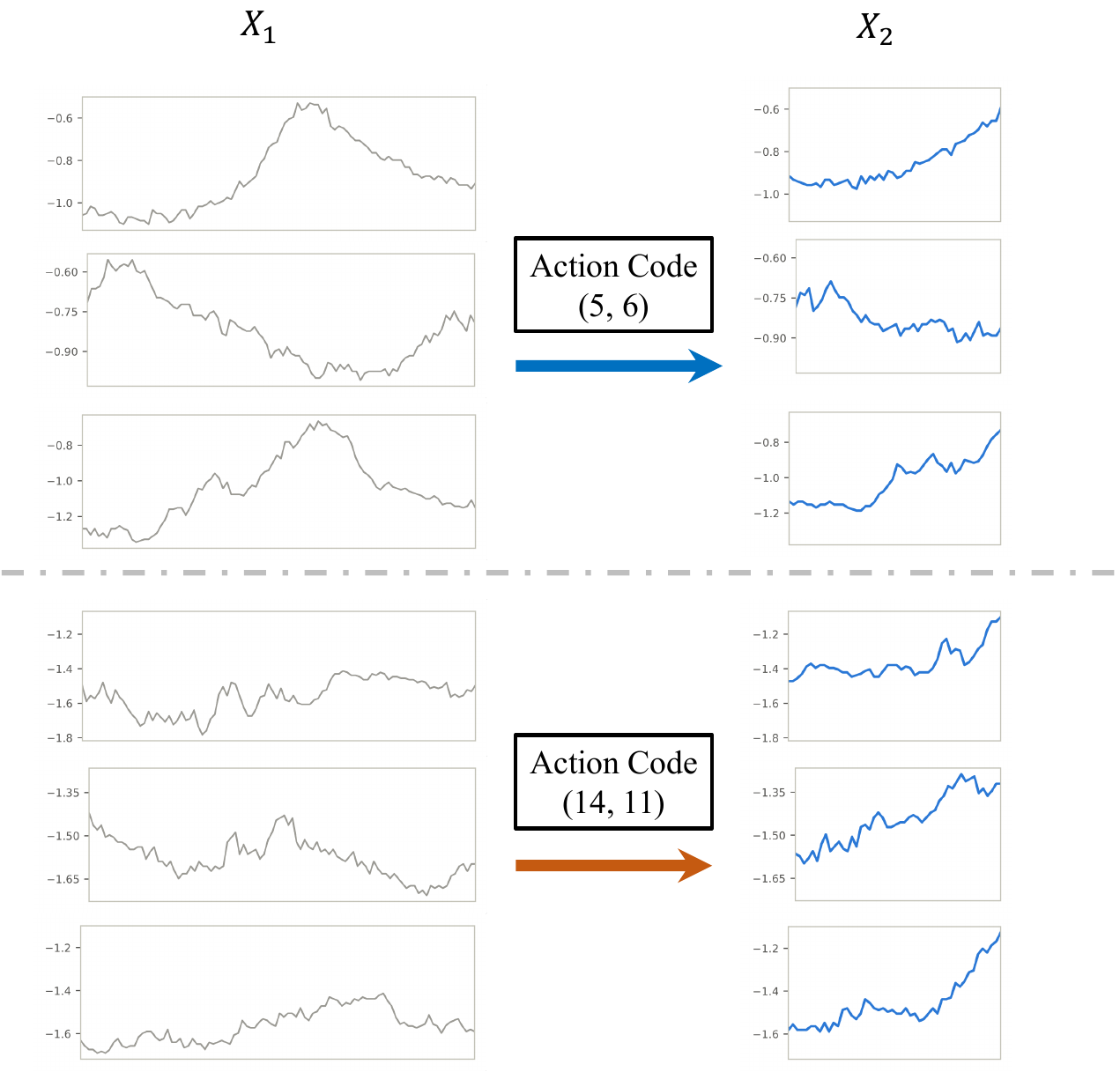}
\caption{Case study of latent actions on univariate forecasting with the JEPA\textsubscript{ACTS} backbone. Contexts $\mathbf{X}_1$ that are assigned the same action code lead to targets $\mathbf{X}_2$ with the same kind of change (horizon 48).}
\label{fig:case_study}
\end{figure}

\subsection{Full Results and Additional Experiments}
\label{appendix:full}

\begin{table*}[p]
\centering
\footnotesize
\setlength{\tabcolsep}{2.5pt}
\renewcommand{\arraystretch}{1.0}
\caption{Time-series classification results of each head averaged across three runs. Datasets marked with $\ddagger$ are not included in pre-training ACTS-WM. The best result is in \textbf{bold}, and $^{\star}$ marks results with ACTS-WM that outperforms original base models. The win rate is the percentage of datasets where ACTS-WM outperforms the base model. Win rates above 50\% are shaded.}
\label{tb:jepa_classify_all_heads_t}
\begin{adjustbox}{max width=\textwidth}
\begin{tabular}{lll*{16}{c}cc}
\toprule
& & & \multicolumn{10}{c}{\textit{Univariate Classification}} & \multicolumn{6}{c}{\textit{Multivariate Classification}} & \multicolumn{2}{c}{Win rate} \\
\cmidrule(lr){4-13} \cmidrule(lr){14-19} \cmidrule(lr){20-21}
\textbf{Head} & \textbf{Model} & & WebT$\ddagger$ & Earthq$\ddagger$ & UWaveX$\ddagger$ & Shapes$\ddagger$ & Worms$\ddagger$ & Comp$\ddagger$ & Insect$\ddagger$ & EOGH$\ddagger$ & Ethanol$\ddagger$ & Semg$\ddagger$ & AWR & BM & CT & FM & SCP2$\ddagger$ & SAD$\ddagger$ & Uni. & Multi. \\
\midrule
 &  & Base & 0.168 & 0.715 & 0.633 & 0.506 & 0.541 & 0.528 & 0.523 & 0.355 & 0.362 & 0.495 & 0.929 & 0.750 & \textbf{0.932} & 0.457 & 0.520 & 0.946 & -- & -- \\
 &  & + \textbf{ACTS-WM} (Pred.) & 0.174$^{\star}$ & \textbf{0.748$^{\star}$} & 0.663$^{\star}$ & 0.538$^{\star}$ & 0.554$^{\star}$ & 0.547$^{\star}$ & 0.460 & 0.328 & 0.250 & 0.413 & \textbf{0.951$^{\star}$} & 0.642 & 0.783 & 0.427 & 0.506 & 0.898 & \cellcolor{oursrowdark}60\% & 17\% \\
 & \multirow{-3}{*}{JEPA$_{\text{ACTS}}$} & + \textbf{ACTS-WM} (Obs.+Pred.) & 0.182$^{\star}$ & 0.739$^{\star}$ & 0.653$^{\star}$ & 0.511$^{\star}$ & 0.558$^{\star}$ & 0.492 & 0.505 & 0.322 & 0.324 & 0.498$^{\star}$ & 0.873 & \textbf{0.858$^{\star}$} & 0.875 & 0.483$^{\star}$ & \textbf{0.522$^{\star}$} & 0.938 & \cellcolor{oursrowdark}60\% & 50\% \\
\rowcolor{gray!10}  &  & Base & 0.195 & \textbf{0.748} & 0.635 & 0.588 & 0.524 & 0.464 & \textbf{0.623} & 0.382 & 0.298 & 0.506 & -- & -- & -- & -- & -- & -- & -- & -- \\
\rowcolor{gray!10}  &  & + \textbf{ACTS-WM} (Pred.) & 0.114 & \textbf{0.748} & \textbf{0.701$^{\star}$} & 0.537 & 0.532$^{\star}$ & 0.623$^{\star}$ & 0.495 & 0.388$^{\star}$ & 0.270 & 0.388 & -- & -- & -- & -- & -- & -- & 40\% & -- \\
\rowcolor{gray!10}  & \multirow{-3}{*}{TS-JEPA} & + \textbf{ACTS-WM} (Obs.+Pred.) & 0.200$^{\star}$ & 0.746 & 0.700$^{\star}$ & 0.560 & 0.511 & 0.500$^{\star}$ & 0.591 & \textbf{0.400$^{\star}$} & 0.268 & \textbf{0.512$^{\star}$} & -- & -- & -- & -- & -- & -- & 50\% & -- \\
 &  & Base & \textbf{0.251} & 0.667 & 0.588 & 0.652 & 0.494 & 0.643 & 0.616 & 0.180 & \textbf{0.457} & 0.331 & 0.948 & 0.467 & 0.912 & 0.500 & 0.500 & 0.946 & -- & -- \\
 &  & + \textbf{ACTS-WM} (Pred.) & 0.181 & 0.736$^{\star}$ & 0.563 & \textbf{0.665$^{\star}$} & \textbf{0.602$^{\star}$} & \textbf{0.673$^{\star}$} & 0.522 & 0.162 & 0.379 & 0.220 & 0.944 & 0.725$^{\star}$ & 0.902 & \textbf{0.507$^{\star}$} & 0.480 & 0.943 & 40\% & 33\% \\
\multirow{-9}{*}{\rotatebox{90}{Linear}} & \multirow{-3}{*}{SC-JEPA} & + \textbf{ACTS-WM} (Obs.+Pred.) & 0.196 & 0.719$^{\star}$ & 0.577 & 0.653$^{\star}$ & 0.571$^{\star}$ & 0.668$^{\star}$ & 0.551 & 0.181$^{\star}$ & 0.377 & 0.239 & 0.949$^{\star}$ & 0.725$^{\star}$ & 0.915$^{\star}$ & 0.503$^{\star}$ & 0.504$^{\star}$ & \textbf{0.954$^{\star}$} & 50\% & \cellcolor{oursrowdark}100\% \\
\midrule
 &  & Base & 0.155 & 0.743 & 0.646 & 0.474 & 0.550 & 0.553 & 0.444 & 0.398 & 0.431 & 0.540 & 0.958 & 0.775 & 0.957 & 0.463 & 0.507 & 0.950 & -- & -- \\
 &  & + \textbf{ACTS-WM} (Pred.) & 0.151 & 0.748$^{\star}$ & 0.665$^{\star}$ & 0.521$^{\star}$ & 0.524 & 0.588$^{\star}$ & 0.494$^{\star}$ & 0.332 & 0.257 & 0.551$^{\star}$ & \textbf{0.962$^{\star}$} & 0.658 & 0.934 & 0.480$^{\star}$ & 0.528$^{\star}$ & 0.562 & \cellcolor{oursrowdark}60\% & 50\% \\
 & \multirow{-3}{*}{JEPA$_{\text{ACTS}}$} & + \textbf{ACTS-WM} (Obs.+Pred.) & 0.173$^{\star}$ & \textbf{0.753$^{\star}$} & 0.663$^{\star}$ & 0.506$^{\star}$ & 0.554$^{\star}$ & 0.545 & 0.478$^{\star}$ & 0.379 & 0.435$^{\star}$ & 0.514 & 0.960$^{\star}$ & \textbf{0.858$^{\star}$} & \textbf{0.961$^{\star}$} & 0.463 & \textbf{0.535$^{\star}$} & 0.948 & \cellcolor{oursrowdark}70\% & \cellcolor{oursrowdark}67\% \\
\rowcolor{gray!10}  &  & Base & 0.229 & 0.748 & 0.697 & 0.652 & 0.550 & 0.508 & \textbf{0.610} & 0.420 & \textbf{0.576} & 0.496 & -- & -- & -- & -- & -- & -- & -- & -- \\
\rowcolor{gray!10}  &  & + \textbf{ACTS-WM} (Pred.) & 0.191 & 0.748 & 0.711$^{\star}$ & 0.655$^{\star}$ & 0.519 & 0.623$^{\star}$ & 0.493 & \textbf{0.437$^{\star}$} & 0.259 & \textbf{0.559$^{\star}$} & -- & -- & -- & -- & -- & -- & 50\% & -- \\
\rowcolor{gray!10}  & \multirow{-3}{*}{TS-JEPA} & + \textbf{ACTS-WM} (Obs.+Pred.) & 0.226 & 0.746 & \textbf{0.718$^{\star}$} & \textbf{0.661$^{\star}$} & \textbf{0.576$^{\star}$} & 0.523$^{\star}$ & 0.609 & 0.424$^{\star}$ & 0.525 & 0.516$^{\star}$ & -- & -- & -- & -- & -- & -- & \cellcolor{oursrowdark}60\% & -- \\
 &  & Base & \textbf{0.265} & 0.707 & 0.591 & 0.622 & 0.498 & 0.660 & 0.599 & 0.159 & 0.447 & 0.326 & 0.937 & 0.492 & 0.942 & \textbf{0.507} & 0.487 & \textbf{0.967} & -- & -- \\
 &  & + \textbf{ACTS-WM} (Pred.) & 0.215 & 0.700 & 0.539 & 0.637$^{\star}$ & 0.563$^{\star}$ & \textbf{0.683$^{\star}$} & 0.473 & 0.140 & 0.363 & 0.197 & 0.926 & 0.792$^{\star}$ & 0.857 & 0.453 & 0.485 & 0.947 & 30\% & 17\% \\
\multirow{-9}{*}{\rotatebox{90}{MLP}} & \multirow{-3}{*}{SC-JEPA} & + \textbf{ACTS-WM} (Obs.+Pred.) & 0.213 & 0.676 & 0.547 & 0.631$^{\star}$ & 0.550$^{\star}$ & 0.673$^{\star}$ & 0.490 & 0.148 & 0.372 & 0.206 & 0.928 & 0.758$^{\star}$ & 0.883 & 0.483 & 0.489$^{\star}$ & 0.952 & 30\% & 33\% \\
\midrule
 &  & Base & 0.152 & \textbf{0.748} & 0.651 & 0.453 & 0.597 & 0.592 & 0.508 & 0.354 & 0.360 & 0.347 & 0.847 & 0.775 & 0.912 & 0.490 & \textbf{0.522} & 0.930 & -- & -- \\
 &  & + \textbf{ACTS-WM} (Pred.) & 0.168$^{\star}$ & \textbf{0.748} & 0.668$^{\star}$ & 0.320 & 0.584 & 0.592 & 0.352 & 0.268 & 0.246 & 0.484$^{\star}$ & 0.897$^{\star}$ & 0.850$^{\star}$ & 0.751 & 0.490 & 0.489 & 0.876 & 30\% & 33\% \\
 & \multirow{-3}{*}{JEPA$_{\text{ACTS}}$} & + \textbf{ACTS-WM} (Obs.+Pred.) & 0.158$^{\star}$ & \textbf{0.748} & 0.668$^{\star}$ & 0.458$^{\star}$ & 0.558 & 0.640$^{\star}$ & 0.490 & 0.348 & 0.396$^{\star}$ & 0.520$^{\star}$ & 0.853$^{\star}$ & \textbf{0.900$^{\star}$} & 0.883 & 0.490 & 0.483 & 0.919 & \cellcolor{oursrowdark}60\% & 33\% \\
\rowcolor{gray!10}  &  & Base & 0.194 & \textbf{0.748} & 0.720 & 0.522 & 0.545 & 0.596 & \textbf{0.627} & \textbf{0.475} & 0.308 & 0.473 & -- & -- & -- & -- & -- & -- & -- & -- \\
\rowcolor{gray!10}  &  & + \textbf{ACTS-WM} (Pred.) & 0.158 & \textbf{0.748} & 0.715 & 0.455 & 0.571$^{\star}$ & 0.636$^{\star}$ & 0.483 & 0.420 & 0.256 & 0.427 & -- & -- & -- & -- & -- & -- & 20\% & -- \\
\rowcolor{gray!10}  & \multirow{-3}{*}{TS-JEPA} & + \textbf{ACTS-WM} (Obs.+Pred.) & 0.202$^{\star}$ & \textbf{0.748} & \textbf{0.727$^{\star}$} & 0.503 & 0.545 & 0.592 & 0.601 & 0.464 & 0.276 & \textbf{0.567$^{\star}$} & -- & -- & -- & -- & -- & -- & 30\% & -- \\
 &  & Base & \textbf{0.268} & \textbf{0.748} & 0.638 & \textbf{0.560} & 0.532 & 0.652 & 0.609 & 0.199 & \textbf{0.480} & 0.333 & \textbf{0.953} & 0.700 & \textbf{0.956} & \textbf{0.520} & 0.494 & \textbf{0.979} & -- & -- \\
 &  & + \textbf{ACTS-WM} (Pred.) & 0.192 & \textbf{0.748} & 0.557 & 0.490 & 0.623$^{\star}$ & \textbf{0.692$^{\star}$} & 0.484 & 0.177 & 0.354 & 0.207 & \textbf{0.953} & 0.800$^{\star}$ & 0.910 & 0.460 & 0.511$^{\star}$ & 0.959 & 20\% & 33\% \\
\multirow{-9}{*}{\rotatebox{90}{SVM}} & \multirow{-3}{*}{SC-JEPA} & + \textbf{ACTS-WM} (Obs.+Pred.) & 0.192 & \textbf{0.748} & 0.558 & 0.490 & \textbf{0.636$^{\star}$} & \textbf{0.692$^{\star}$} & 0.489 & 0.171 & 0.356 & 0.209 & \textbf{0.953} & 0.800$^{\star}$ & 0.914 & 0.480 & 0.489 & 0.960 & 20\% & 17\% \\
\bottomrule[1.3pt]
\end{tabular}
\end{adjustbox}
\vspace{6pt}
\caption{Time-series forecasting MSE ($H=48$) of each head averaged across three runs. Datasets marked with $\ddagger$ are not included in pre-training ACTS-WM. The best result is in \textbf{bold}, and $^{\star}$ marks results with ACTS-WM that outperforms original base models. The win rate is the percentage of datasets where ACTS-WM outperforms the base model. Win rates above 50\% are shaded.}
\label{tb:jepa_forecast_mse_all_heads_t}
\begin{adjustbox}{max width=\textwidth}
\begin{tabular}{lll*{11}{c}cc}
\toprule
& & & \multicolumn{6}{c}{\textit{Univariate Forecasting}} & \multicolumn{5}{c}{\textit{Multivariate Forecasting}} & \multicolumn{2}{c}{Win rate} \\
\cmidrule(lr){4-9} \cmidrule(lr){10-14} \cmidrule(lr){15-16}
\textbf{Head} & \textbf{Model} & & Wiki$\ddagger$ & Weather & ETTh1 & ETTh2 & ETTm1 & ETTm2 & Weather & ETTh1 & ETTh2 & ETTm1 & ETTm2 & Uni. & Multi. \\
\midrule
 &  & Base & 1.167 & 0.007 & 0.100 & 0.478 & 0.044 & 0.184 & 0.257 & 1.056 & 0.302 & 0.597 & 0.161 & -- & -- \\
 &  & + \textbf{ACTS-WM} (Pred.) & 1.063$^{\star}$ & 0.003$^{\star}$ & \textbf{0.084$^{\star}$} & \textbf{0.351$^{\star}$} & \textbf{0.035$^{\star}$} & 0.138$^{\star}$ & \textbf{0.238$^{\star}$} & \textbf{0.822$^{\star}$} & \textbf{0.264$^{\star}$} & \textbf{0.489$^{\star}$} & \textbf{0.152$^{\star}$} & \cellcolor{oursrowdark}100\% & \cellcolor{oursrowdark}100\% \\
 & \multirow{-3}{*}{JEPA$_{\text{ACTS}}$} & + \textbf{ACTS-WM} (Obs.+Pred.) & 1.164$^{\star}$ & 0.007 & 0.086$^{\star}$ & 0.444$^{\star}$ & 0.043$^{\star}$ & 0.175$^{\star}$ & 0.259 & 1.017$^{\star}$ & 0.290$^{\star}$ & 0.564$^{\star}$ & 0.156$^{\star}$ & \cellcolor{oursrowdark}83\% & \cellcolor{oursrowdark}80\% \\
\rowcolor{gray!10}  &  & Base & 1.047 & 0.002 & 0.085 & 0.402 & 0.036 & 0.153 & -- & -- & -- & -- & -- & -- & -- \\
\rowcolor{gray!10}  &  & + \textbf{ACTS-WM} (Pred.) & \textbf{0.972$^{\star}$} & \textbf{0.001$^{\star}$} & 0.097 & \textbf{0.351$^{\star}$} & 0.064 & 0.426 & -- & -- & -- & -- & -- & 50\% & -- \\
\rowcolor{gray!10}  & \multirow{-3}{*}{TS-JEPA} & + \textbf{ACTS-WM} (Obs.+Pred.) & 1.040$^{\star}$ & 0.002 & 0.088 & 0.396$^{\star}$ & 0.045 & \textbf{0.136$^{\star}$} & -- & -- & -- & -- & -- & 50\% & -- \\
 &  & Base & 1.225 & 0.003 & 0.740 & 1.977 & 1.312 & 1.602 & 0.590 & 1.207 & 2.164 & 1.163 & 0.914 & -- & -- \\
 &  & + \textbf{ACTS-WM} (Pred.) & 1.288 & 0.003 & 0.428$^{\star}$ & 6.351 & 0.793$^{\star}$ & 1.673 & 0.471$^{\star}$ & 1.279 & 2.213 & 1.255 & 0.909$^{\star}$ & 33\% & 40\% \\
\multirow{-9}{*}{\rotatebox{90}{Linear}} & \multirow{-3}{*}{SC-JEPA} & + \textbf{ACTS-WM} (Obs.+Pred.) & 1.240 & 0.003 & 1.089 & 6.716 & 0.775$^{\star}$ & 1.073$^{\star}$ & 0.514$^{\star}$ & 1.217 & 2.261 & 1.142$^{\star}$ & 0.817$^{\star}$ & 33\% & \cellcolor{oursrowdark}60\% \\
\midrule
 &  & Base & 1.141 & \textbf{0.005} & 0.095 & 0.467 & 0.049 & 0.178 & 0.255 & 1.218 & 0.366 & 0.605 & 0.160 & -- & -- \\
 &  & + \textbf{ACTS-WM} (Pred.) & 1.092$^{\star}$ & \textbf{0.005} & 0.094$^{\star}$ & 0.345$^{\star}$ & \textbf{0.045$^{\star}$} & 0.139$^{\star}$ & 1.902 & \textbf{0.678$^{\star}$} & \textbf{0.276$^{\star}$} & \textbf{0.495$^{\star}$} & \textbf{0.153$^{\star}$} & \cellcolor{oursrowdark}83\% & \cellcolor{oursrowdark}80\% \\
 & \multirow{-3}{*}{JEPA$_{\text{ACTS}}$} & + \textbf{ACTS-WM} (Obs.+Pred.) & 1.141 & \textbf{0.005} & 0.093$^{\star}$ & 0.468 & 0.049 & 0.171$^{\star}$ & \textbf{0.253$^{\star}$} & 1.196$^{\star}$ & 0.297$^{\star}$ & 0.583$^{\star}$ & 0.156$^{\star}$ & 33\% & \cellcolor{oursrowdark}100\% \\
\rowcolor{gray!10}  &  & Base & 0.976 & \textbf{0.005} & 0.097 & \textbf{0.319} & \textbf{0.045} & 0.197 & -- & -- & -- & -- & -- & -- & -- \\
\rowcolor{gray!10}  &  & + \textbf{ACTS-WM} (Pred.) & 0.977 & \textbf{0.005} & 0.157 & 0.412 & \textbf{0.045} & 0.184$^{\star}$ & -- & -- & -- & -- & -- & 17\% & -- \\
\rowcolor{gray!10}  & \multirow{-3}{*}{TS-JEPA} & + \textbf{ACTS-WM} (Obs.+Pred.) & \textbf{0.968$^{\star}$} & \textbf{0.005} & \textbf{0.091$^{\star}$} & 0.322 & 0.054 & \textbf{0.133$^{\star}$} & -- & -- & -- & -- & -- & 50\% & -- \\
 &  & Base & 1.239 & 0.006 & 0.601 & 1.024 & 1.448 & 1.543 & 0.558 & 1.194 & 1.919 & 1.217 & 0.915 & -- & -- \\
 &  & + \textbf{ACTS-WM} (Pred.) & 1.270 & 0.006 & 0.946 & 1.448 & 0.964$^{\star}$ & 1.651 & 0.483$^{\star}$ & 1.201 & 2.050 & 1.269 & 0.981 & 17\% & 20\% \\
\multirow{-9}{*}{\rotatebox{90}{Ridge}} & \multirow{-3}{*}{SC-JEPA} & + \textbf{ACTS-WM} (Obs.+Pred.) & 1.272 & 0.006 & 0.946 & 1.448 & 0.915$^{\star}$ & 1.614 & 0.491$^{\star}$ & 1.201 & 2.051 & 1.268 & 0.919 & 17\% & 20\% \\
\midrule
 &  & Base & 1.268 & \textbf{0.016} & \textbf{0.104} & \textbf{0.278} & \textbf{0.077} & \textbf{0.186} & \textbf{0.278} & \textbf{0.573} & \textbf{0.233} & \textbf{0.542} & \textbf{0.162} & -- & -- \\
 &  & + \textbf{ACTS-WM} (Pred.) & 1.393 & 0.042 & 0.671 & 0.712 & 0.638 & 0.665 & 0.554 & 1.084 & 1.136 & 1.018 & 1.096 & 0\% & 0\% \\
 & \multirow{-3}{*}{JEPA$_{\text{ACTS}}$} & + \textbf{ACTS-WM} (Obs.+Pred.) & 1.496 & 0.088 & 0.499 & 0.559 & 0.485 & 0.506 & 0.474 & 0.888 & 0.809 & 0.861 & 0.784 & 0\% & 0\% \\
\rowcolor{gray!10}  &  & Base & \textbf{1.151} & 0.543 & 0.215 & 0.674 & 0.144 & 0.397 & -- & -- & -- & -- & -- & -- & -- \\
\rowcolor{gray!10}  &  & + \textbf{ACTS-WM} (Pred.) & 1.372 & 0.099$^{\star}$ & 0.822 & 0.815 & 0.372 & 0.528 & -- & -- & -- & -- & -- & 17\% & -- \\
\rowcolor{gray!10}  & \multirow{-3}{*}{TS-JEPA} & + \textbf{ACTS-WM} (Obs.+Pred.) & 1.976 & 6.494 & 0.519 & 1.407 & 1.273 & 2.899 & -- & -- & -- & -- & -- & 0\% & -- \\
 &  & Base & 1.413 & 0.148 & 1.631 & 1.478 & 1.822 & 1.625 & 0.848 & 1.321 & 1.890 & 1.434 & 1.913 & -- & -- \\
 &  & + \textbf{ACTS-WM} (Pred.) & 1.363$^{\star}$ & 0.018$^{\star}$ & 0.728$^{\star}$ & 1.358$^{\star}$ & 1.588$^{\star}$ & 1.354$^{\star}$ & 0.641$^{\star}$ & 1.129$^{\star}$ & 1.395$^{\star}$ & 1.105$^{\star}$ & 1.596$^{\star}$ & \cellcolor{oursrowdark}100\% & \cellcolor{oursrowdark}100\% \\
\multirow{-9}{*}{\rotatebox{90}{NLH}} & \multirow{-3}{*}{SC-JEPA} & + \textbf{ACTS-WM} (Obs.+Pred.) & 1.906 & 0.473 & 1.739 & 2.140 & 1.493$^{\star}$ & 4.001 & 1.009 & 1.209$^{\star}$ & 2.084 & 1.599 & 1.740$^{\star}$ & 17\% & 40\% \\
\bottomrule[1.3pt]
\end{tabular}
\end{adjustbox}
\end{table*}

\subsubsection{Performance over Downstream Heads}

The results of each downstream head are reported in Table~\ref{tb:jepa_classify_all_heads_t} and Table~\ref{tb:jepa_forecast_mse_all_heads_t}.

For forecasting, the improvements brought by ACTS-WM are mainly observed with the linear head and ridge head. With these two heads, ACTS-WM outperforms JEPA$_\text{{ACTS}}$ base model on most datasets. Remarkably, the linear head outperforms the base model on all datasets when JEPA$_\text{{ACTS}}$ is integrated with ACTS-WM in the predictor-only setting. Although ACTS-WM does not yield substantial improvements with NLH head for JEPA$_\text{{ACTS}}$, the NLH head demonstrates its effectiveness with SC-JEPA, which outperforms the base in all datasets.

For classification, JEPA$_\text{{ACTS}}$ outperforms the base model on most datasets in univariate classification. For multivariate classification, SC-JEPA with ACTS-WM outperforms the base model on all datasets when both observation and predictor embeddings are fed into the linear head.

\subsubsection{Case study of latent action semantics}\label{subsec:case_study}

In Figure~\ref{fig:case_study}, we plot the time-series pairs associated with different action codes. 

For the action code with index $(5, 6)$, the $X_2$ sequences of selected pairs exhibit patterns similar to those in the middle sections of their corresponding $X_1$ sequences. For the action code with index $(14, 11)$, the $X_2$ sequences of all three pairs exhibit an upward trend regardless of whether the tail sections of $X_1$ exhibit an upward trend, a downward trend or a steady state.

Based on these observations, we speculate that the actions may capture some meaningful patterns that can be interpreted in terms of human-understandable concepts.

\end{document}